%% file: main.tex
\documentclass[runningheads]{llncs}

\usepackage{eccv}

\usepackage{eccvabbrv}  
\usepackage{graphicx}
\usepackage{booktabs}
\usepackage{xcolor}
\usepackage{framed}
\usepackage{tcolorbox}
\usepackage[accsupp]{axessibility}  
\usepackage{xcolor}

\usepackage{algorithm}
\usepackage{algpseudocode}
\usepackage{booktabs}
\usepackage{multirow}
\usepackage{dblfloatfix}
\usepackage{colortbl}
\usepackage{xcolor}

\usepackage{caption}
\usepackage{tabularx}
\usepackage{array}
\usepackage{colortbl}
\usepackage{makecell}
\usepackage{pifont} 
\newcolumntype{L}[1]{>{\raggedright\arraybackslash}p{#1}}
\newcolumntype{C}[1]{>{\centering\arraybackslash}p{#1}}
\newcolumntype{Y}{>{\raggedright\arraybackslash}X}
\definecolor{lightgray}{gray}{0.93}
\definecolor{headergray}{gray}{0.88}

\usepackage{hyperref}
\usepackage{orcidlink}
\usepackage[dvipsnames]{xcolor}

\usepackage{algorithm}
\usepackage{algpseudocode}
\usepackage{graphicx}
\usepackage{tikz}
\usepackage{pgfplots}
\pgfplotsset{compat=1.17}
\usetikzlibrary{positioning, shapes, arrows.meta}
\usepackage[accsupp]{axessibility}  

\usetikzlibrary{positioning,fit}
\usepackage[misc]{ifsym} 

\begin{document}

\title{Seeing Through the Weights: Privacy Leakage in Scene Coordinate Regression}

\titlerunning{Seeing Through the Weights}


\newcommand{\equalcontrib}{\textsuperscript{*}}
\definecolor{raspberry}{RGB}{199, 21, 133}

\newcommand{\corrauth}{\ensuremath{^\dagger}}

\author{Oleksii Nasypanyi\equalcontrib\inst{1}\orcidlink{0009-0009-5353-6689} \and
Jaemin Cho\equalcontrib\inst{1}\orcidlink{0009-0003-8984-3505} \and
Utku Ozbulak\inst{2,3}\orcidlink{0000-0003-3084-6034} \and
Byungkon Kang\inst{4}\orcidlink{0000-0001-8541-2861} \and 
Francois Rameau\corrauth\inst{4}\orcidlink{0000-0001-5031-7653}}

\authorrunning{O. Nasypanyi et al. }

\institute{
\textsuperscript{1}Stony Brook University \quad
\textsuperscript{2}Ghent University Global Campus \quad \\
\textsuperscript{3}George Mason University Korea  \quad
\textsuperscript{4}SUNY Korea
}

\maketitle

\begingroup
\renewcommand{\thefootnote}{}
\footnotetext{
\ensuremath{^*}Equal contribution: \{oleksii.nasypanyi, jaemin.cho\}@stonybrook.edu\\
\ensuremath{^\dagger}Corresponding author: \href{mailto:francois.rameau@sunykorea.ac.kr}{francois.rameau@sunykorea.ac.kr}
}
\endgroup

\begin{abstract}
  \input{sec/0-abstract}

\end{abstract}

\input{sec/1-intro}

\input{sec/2-litterature}

\input{sec/3-background}

\input{sec/4-Methodology}
\input{sec/5-experiments}
\input{sec/6-conclusion}

%
%
\bibliographystyle{splncs04}
\bibliography{main}

\clearpage

\input{supplementary/supl_arxiv}

\end{document}

%% file: sec/0-abstract.tex
Scene Coordinate Regression (SCR) methods are increasingly adopted for visual localization. In these approaches, the scene is implicitly encoded within a neural network that regresses a 3D world coordinate for each image pixel. Because the scene is represented only through the network parameters and not stored explicitly as images or maps, such methods are often assumed to be privacy-preserving. In this work, we show that this assumption is incorrect in practice.

Specifically, we introduce a query-based attack that reconstructs the 3D geometry of the training environment from an SCR model under different levels of model access.
To do so, we repeatedly query the model with batches of proxy images unrelated to the target scene to obtain dense pixel-wise 3D coordinates. 
Reliable points are identified through their stability under small input perturbations and can be further refined in a white-box setting. These stable points are accumulated across independent query batches to recover the scene geometry. From the recovered 3D representation, we also invert the network features to synthesize images from arbitrary viewpoints, revealing additional appearance information.

Experiments on indoor and outdoor datasets demonstrate that substantial portions of training environments can be reconstructed with high geometric fidelity. Beyond geometry, we also recover an approximate color appearance, which exposes recognizable layout and potentially sensitive scene elements.
This directly contradicts claims in the literature that SCR representations are privacy-preserving by design, and reveals a real risk when such systems are deployed in private or security-critical spaces.
The project page is available 
\href{https://jaeminch0.github.io/seeing-through-the-weights-privacy-leakage-in-scene-coordinate-regression/}
{\textcolor{raspberry}{\textit{here}}}.

\keywords{Visual Localization \and Privacy Leakage \and Model Inversion}

%% file: sec/1-intro.tex
\section{Introduction}
\label{sec:intro}

\begin{figure}[t]
    \centering
    \includegraphics[width=0.9\linewidth]{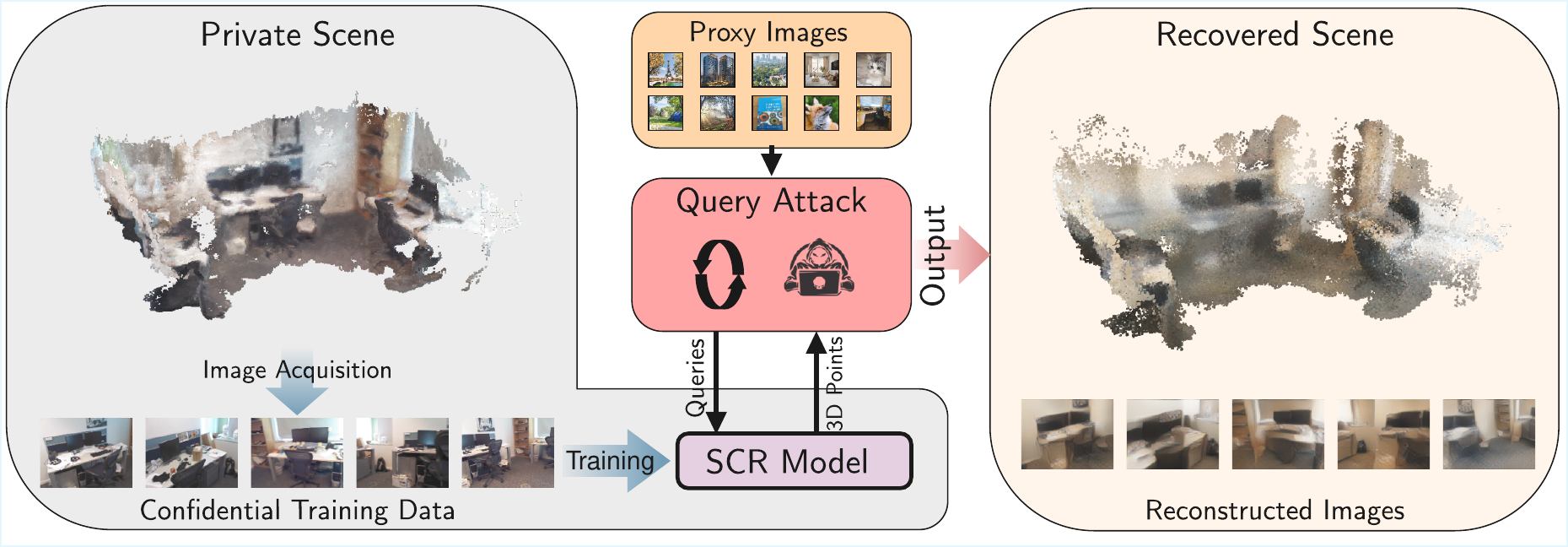}
\caption{ Overview of the proposed query-based attack.
An SCR model trained on images of a private scene is queried using proxy images. By aggregating the returned 3D predictions, the attacker reconstructs a voxelized representation of the scene and synthesizes novel views, revealing both geometry and appearance. \vspace{-0.5cm}}
    \label{fig:teaser}    
\end{figure}

Given an image of a known environment, visual localization aims to recover the camera’s six-degree-of-freedom pose. It finds applications in augmented reality~\cite{sarlin2022lamar}, autonomous navigation~\cite{jeong2024multi}, and robotics~\cite{lee2021large}.
As these systems are increasingly used in private or sensitive spaces like homes, offices, or medical facilities, privacy becomes a real concern~\cite{chelani2025privacy}.
In such contexts, visual localization systems may inadvertently expose information about the environment for which they were built. This leakage may reveal structural clues such as scene layout or geometry, or even appearance information, resulting in a privacy breach.
This issue is particularly obvious for hierarchical localization techniques~\cite{pittaluga2019revealing} based on Structure-from-Motion (SfM) maps, where the 3D structure of the scene is directly and explicitly available. Prior works have addressed this by replacing 3D points with lines~\cite{speciale2019privacy} to prevent model inversion via feature reprojection~\cite{dosovitskiy2016inverting}. 
However, this problem is far less explored for recent learning-based approaches such as Scene Coordinate Regression (SCR)~\cite{brachmann2017dsac, brachmann2023accelerated}, which represent the scene implicitly through a neural network that predicts per-pixel 3D coordinates in a fixed world coordinate system. 
Because the scene is not stored explicitly as images or maps and exists only through the network parameters, these methods are widely believed to be inherently privacy-preserving.
Such claims appear repeatedly in the literature~\cite{chelani2025obfuscation,Do_2022_SceneLandmarkLoc,zhou2022geometry,brachmann2023accelerated}. For example:
 \vspace{-0.1cm}
\begin{center}
\fcolorbox{black!15}{black!5}{%
  \begin{minipage}{0.95\linewidth}
 \fontsize{8}{5}\selectfont
  \itshape

  ``The resulting maps can be as small as 4MB, and privacy preserving''
  \hfill {\normalfont\cite{brachmann2023accelerated}}

  \vspace{0.4em}

``SCR is said to be inherently privacy-preserving because there is no set of 2D or 3D points to run the inversion attack on.''
  \hfill {\normalfont\cite{chelani2025obfuscation}}

  \vspace{0.4em}

    ``These methods implicitly encode scene information in the learned parameters of a convolutional neural network (CNN), rather than explicitly storing images or features. Thus, they preserve privacy by design.''
  \hfill {\normalfont\cite{Do_2022_SceneLandmarkLoc}}

  \vspace{0.4em}

    ``DSAC++ is the most accurate method while being resilient to privacy attacks as it does not need to transmit visual descriptors''
  \hfill {\normalfont\cite{zhou2022geometry}}

  \vspace{0.4em}
  
  \end{minipage}%
}
\end{center}

In this work, we show that this assumption does not hold in practice. We demonstrate that the structure and appearance of a training scene can be extracted from an SCR network through a query-based attack.
For this purpose, we consider an adversary with query access to a trained SCR model under black-box, gray-box, or white-box settings, reflecting increasing levels of access to the feature extractor and model parameters. In all cases, the adversary has no access to the training images or explicit scene geometry. Despite these constraints, our attack recovers detailed information about the scene, as depicted in~\figurename~\ref{fig:teaser}.

Our attack consists of multiple stages.
We begin by querying the SCR network with batches of proxy images drawn from natural image datasets unrelated to the target scene, obtaining dense pixel-wise 3D coordinate predictions. While individual predictions are noisy, aggregating them across many batches already reveals a coarse structure of the scene. 
This alone constitutes a black-box attack requiring no access to model internals.

This approach can be significantly improved when access to the feature extractor is available. SCR models are well-behaved near features observed during training but become unstable outside that distribution~\cite{novak2018sensitivity}. Since proxy images are unrelated to the target scene, the features they induce frequently fall into these unstable regions, where small perturbations cause large changes in predicted 3D coordinates. We exploit this property in two ways: first, stability under feature perturbations serves as a criterion to identify and discard unreliable predictions; second, in the white-box setting, features can be refined by gradient descent to move them toward stable regions corresponding to geometry encoded during training.

The remaining predictions are accumulated across proxy batches using a voxel grid. Since all predictions share the same learned coordinate system, voxels consistently populated across batches correspond to true scene geometry.

Finally, recovered feature vectors are projected onto arbitrary virtual camera poses and inverted back to image space via a feature inversion network, revealing approximate scene appearance.

\noindent This paper makes the following contributions:
(i) we provide, to the best of our knowledge, the first systematic evidence that SCR models are not inherently privacy-preserving;
(ii) we present the first   attack framework that reconstructs scene geometry from trained SCR models under different access levels;
(iii) we further recover coarse appearance by inverting reconstructed features, allowing recognizable view synthesis from novel viewpoints;
(iv) we validate the attack on standard benchmarks, recovering substantial portions of the training environments.

%% file: sec/2-litterature.tex
\section{Related Works}

In this section, we will review the different visual localization approaches, their vulnerabilities and the solutions that have been deployed for their defense.

\vspace{0.2em}
\noindent\textbf{Visual Localization}
\vspace{0.2em}

\noindent Recent visual localization approaches fall into 3 main families: hierarchical localization, absolute pose regression (APR), and scene coordinate regression (SCR).

Hierarchical localization estimates the 6DoF camera pose within a pre-built SfM map using a coarse-to-fine pipeline. It typically begins with global image retrieval~\cite{arandjelovic2016netvlad, izquierdo2024optimal}, followed by local feature matching~\cite{lowe2004distinctive,detone2018superpoint} against the 3D map, and final pose estimation via PnP-RANSAC~\cite{larsson2017making}. Although highly accurate and scalable, these methods require storing large maps, incur significant computational cost, and do not inherently protect the privacy of the mapping images~\cite{pittaluga2019revealing}.

To mitigate these limitations, APR methods~\cite{kendall2015posenet, clark2017vidloc, clark2017vinet, song2025method} directly regress camera pose from a single image, implicitly encoding scene information in network parameters. However, APR suffers from limited accuracy and scalability. 

SCR improves accuracy by predicting a 3D world coordinate for each pixel, yielding dense 2D–3D correspondences for PnP-RANSAC. Originally introduced with random forests~\cite{shotton2013scene} and later extended with differentiable RANSAC~\cite{brachmann2017dsac}, SCR achieves state-of-the-art accuracy on small scenes but requires long training times. Accelerated Coordinate Encoding (ACE) decouples SCR into a frozen scene-agnostic backbone and a lightweight scene-specific MLP~\cite{brachmann2023accelerated}, reducing training from hours to minutes, with extensions to uncalibrated cameras~\cite{brachmann2024scene}, large-scale environments~\cite{wang2024glace}, and improved generalization~\cite{bruns2025ace}. Beyond performance, SCR methods are often regarded as privacy-preserving; an assumption we challenge in this work.

\vspace{0.2em}
\noindent\textbf{Privacy in Visual Localization}
\vspace{0.2em}

\noindent 
Among the first to highlight model inversion risks in visual localization, \cite{pittaluga2019revealing} showed that images can be reconstructed from sparse SfM maps via feature reprojection and inversion.
To counter such attacks, Speciale et al.~\cite{speciale2019privacy} proposed replacing 3D points with randomly oriented lines, preserving enough geometry for pose estimation~\cite{zhou2020complete} while theoretically preventing map inversion. This was extended to protect client-side 2D keypoints~\cite{speciale2019privacy-query} and later to privacy-preserving SfM~\cite{geppert2020privacy}.
Despite the popularity of such line-based defenses, they have proven not to be entirely robust. As shown in~\cite{chelani2021privacy}, the original 3D points can be recovered from line clouds using density-based statistics. This vulnerability motivated the development of more sophisticated lifting defense strategies, including Paired-Point Lifting~\cite{lee2023paired}, 3D Ray Clouds~\cite{moon2024efficient}, and Coordinate Permutations~\cite{pan2023privacy}.
However, these advanced representations were shown to remain vulnerable to attacks exploiting neighborhood information encoded in feature descriptors~\cite{chelani2025obfuscation}.

In parallel to geometric defenses, another research direction focuses on feature-level privacy to prevent scene reconstruction.  For instance, \cite{dusmanu2021privacy} and~\cite{ng2022ninjadesc} utilize adversarial learning to conceal sensitive content within the descriptors, while~\cite{pittaluga2023ldp} and~\cite{pietrantoni2023segloc} provide stronger protection by introducing differential privacy or leveraging abstract semantic labels that are inherently difficult to invert.

Recent work has attempted to improve privacy in NeRF~\cite{mildenhall2021nerf} and Gaussian Splatting~\cite{kerbl20233d} by removing RGB supervision~\cite{pietrantoni2025can,Pietrantoni_2025_CVPR}, but this only masks visual appearance, while the scene structure and its content are still directly recoverable.
In contrast, SCR models~\cite{brachmann2017dsac, brachmann2023accelerated} are never trained to render or reconstruct visual appearance; their objective is strictly to map pixels to 3D coordinates. As a result, they are commonly viewed as privacy-preserving~\cite{Do_2022_SceneLandmarkLoc,brachmann2023accelerated,zhou2022geometry}. 
This assumption has not been rigorously evaluated and appears unfounded. 
Unlike prior inversion methods~\cite{chelani2025privacy} that operate on features extracted from target scene images or explicit 3D maps, our attack has access to neither: it relies solely on features from unrelated proxy images, with no knowledge of the scene geometry. Despite this constrained setting, we show that substantial scene information can still be recovered.

%% file: sec/3-background.tex
\begin{figure}[t]
    \centering
    \includegraphics[width=0.8\linewidth]{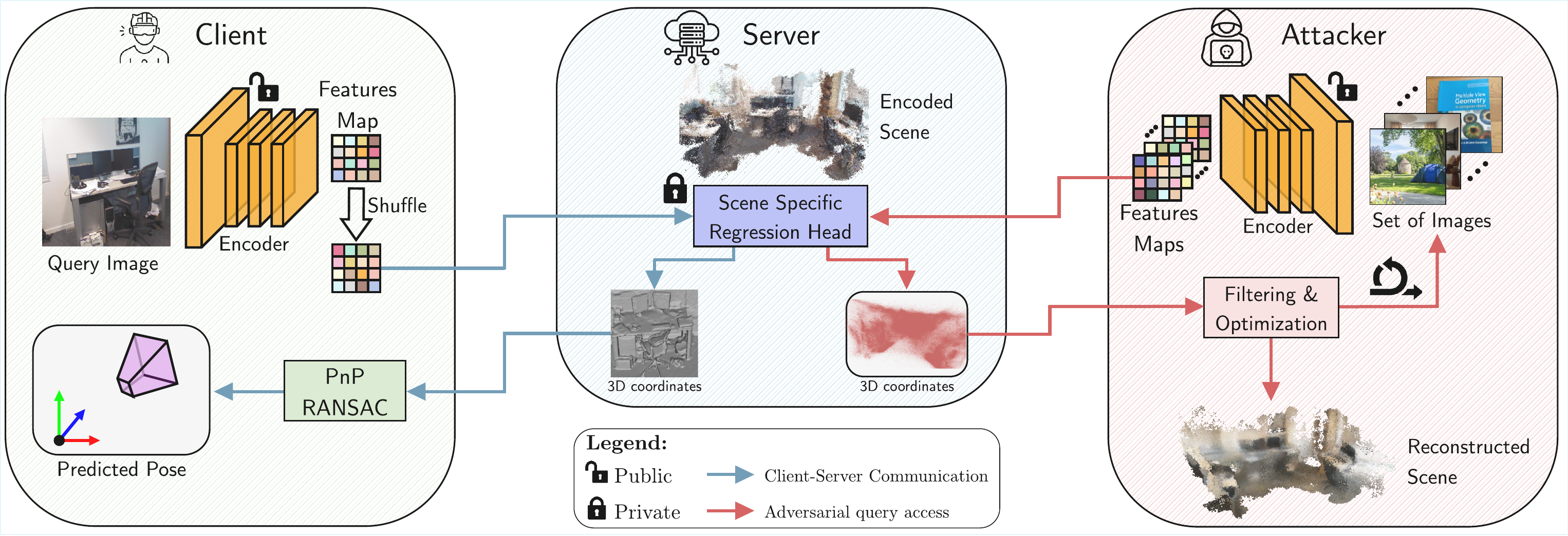}
\caption{Gray-box deployment scenario considered in this work. Feature extraction is performed on the client using a public encoder, while the scene-specific head is hosted remotely. The client transmits spatially shuffled feature maps to the server, which returns predicted 3D scene coordinates used locally for pose estimation. \vspace{-0.5cm}}
    \label{fig:atack_scenario}    
\end{figure}

\vspace{-0.2cm}
\section{Background}
\label{Sec::Background}
\vspace{-0.3cm}
This section introduces SCR, the ACE framework as a representative SCR example used in this work, and the deployment settings considered in our analysis.

\subsection{Scene Coordinate Regression and ACE}

Scene Coordinate Regression formulates visual localization by predicting dense 2D–3D correspondences with a neural network, rather than regressing the camera pose directly. 
Given an input image $I \in \mathbb{R}^{C_{\text{img}} \times H \times W}$ (where $C_{\text{img}}=1$ for grayscale and $C_{\text{img}}=3$ for RGB), an SCR model parameterized by $\theta$ predicts a dense set of 3D scene coordinates $\mathbf{X} = h_\theta(I) = \{\mathbf{x}_{1}, \dots, \mathbf{x}_{HW}\}$, where each $\mathbf{x}_i \in \mathbb{R}^3$ corresponds to the 3D scene coordinate predicted for one image pixel, expressed in a fixed world coordinate system. These correspondences are then used within a PnP–RANSAC pipeline to estimate the 6DoF camera pose.

While our attack is compatible with most SCR models, we use ACE~\cite{brachmann2023accelerated} as our main illustrative case. This choice is motivated by its efficiency and explicit separation between scene-agnostic and scene-specific components, which is well-suited to server-client deployments commonly used in privacy-preserving localization pipelines~\cite{chelani2023privacy}.  
As mentioned earlier, ACE models $h_\theta$ as a two-stage process.
A pretrained, scene-agnostic CNN backbone $\phi$ maps an input grayscale image $I$ to a dense feature map $\mathbf{F} = \phi(I) \in \mathbb{R}^{C \times H' \times W'}$,  where $C$ is the feature channel dimension and $(H', W')$ is the spatial resolution after downsampling, from which a scene-specific MLP $g_{\theta_s}$ predicts a dense field of 3D scene coordinates  $\mathbf{X} = g_{\theta_s}(\mathbf{F}) \in \mathbb{R}^{3 \times H' \times W'}$.

\subsection{Threat Model}
\label{sec:threat-model}

We adopt standard adversarial access models commonly used in machine learning security and model extraction settings~\cite{athalye2018obfuscated,carlini2017towards}, and analyze SCR systems under three levels of attacker access. These settings correspond to increasing visibility into the model pipeline, ranging from query-only access to partial or full access to model components.

\noindent\textbf{Black-box access:} 
The localization pipeline is hosted remotely and exposed through a prediction API where clients transmit input images to the server, which returns predicted 3D point maps or camera poses. In a standard black-box inference interface, the attacker can issue arbitrary queries and observe model outputs, while having no access to intermediate representations or model parameters~\cite{tramer2016stealing}. This setting corresponds to the weakest access model, where the attacker can only interact with the system through input-output queries and has no visibility into intermediate representations or model parameters. 

\noindent\textbf{Gray-box access:} 
In this setting, feature extraction is performed on the client using a public, scene-agnostic encoder $\phi$, while the scene-specific regression head $g_{\theta_s}$ is hosted remotely and accessed via an API that accepts feature maps and returns predicted 3D coordinates. This split architecture is commonly considered in client--server localization systems~\cite{chelani2025obfuscation}. Since SCR feature maps are dense and may retain appearance information, they can be vulnerable to inversion-based reconstruction~\cite{pittaluga2019revealing}. As a simple privacy-preserving design choice, we consider a variant in which the client applies a random spatial permutation to the feature map prior to transmission, removing explicit 2D spatial ordering while preserving per-feature descriptors. The server therefore operates on unordered features and returns predicted 3D coordinates, while pose estimation is performed on the client, as illustrated in \figurename~\ref{fig:atack_scenario}.

\noindent\textbf{White-box access:} 
In this setting, the attacker has access to the parameters of $g_{\theta_s}$ and can compute gradients through the model. This may occur, for example, when scene-specific heads are deployed on-device or when model parameters are otherwise exposed. While such exposure is not typical in standard deployments, it can arise when models are assumed to be inherently privacy-preserving and thus distributed without additional safeguards. This setting therefore represents a worst-case scenario in terms of potential information leakage under maximal adversarial access.%

\noindent\textbf{Objective of the adversary:}
Across all settings, the attacker seeks to reconstruct the private training scene geometry $\mathcal{G} = \{\mathbf{x}_i\}_{i=1}^{N}$ using the capabilities permitted by the given access model (e.g., query access, intermediate features, or gradients) together with a collection of proxy images unrelated to the target scene. Importantly, the evaluated threat models assume no access to the original training images or ground-truth scene geometry.

%% file: sec/4-Methodology.tex
\vspace{-0.2cm}
\section{Method}
\vspace{-0.1cm}
\label{fig:method}

\begin{figure}[t]
    \centering
    \includegraphics[width=0.9\linewidth]{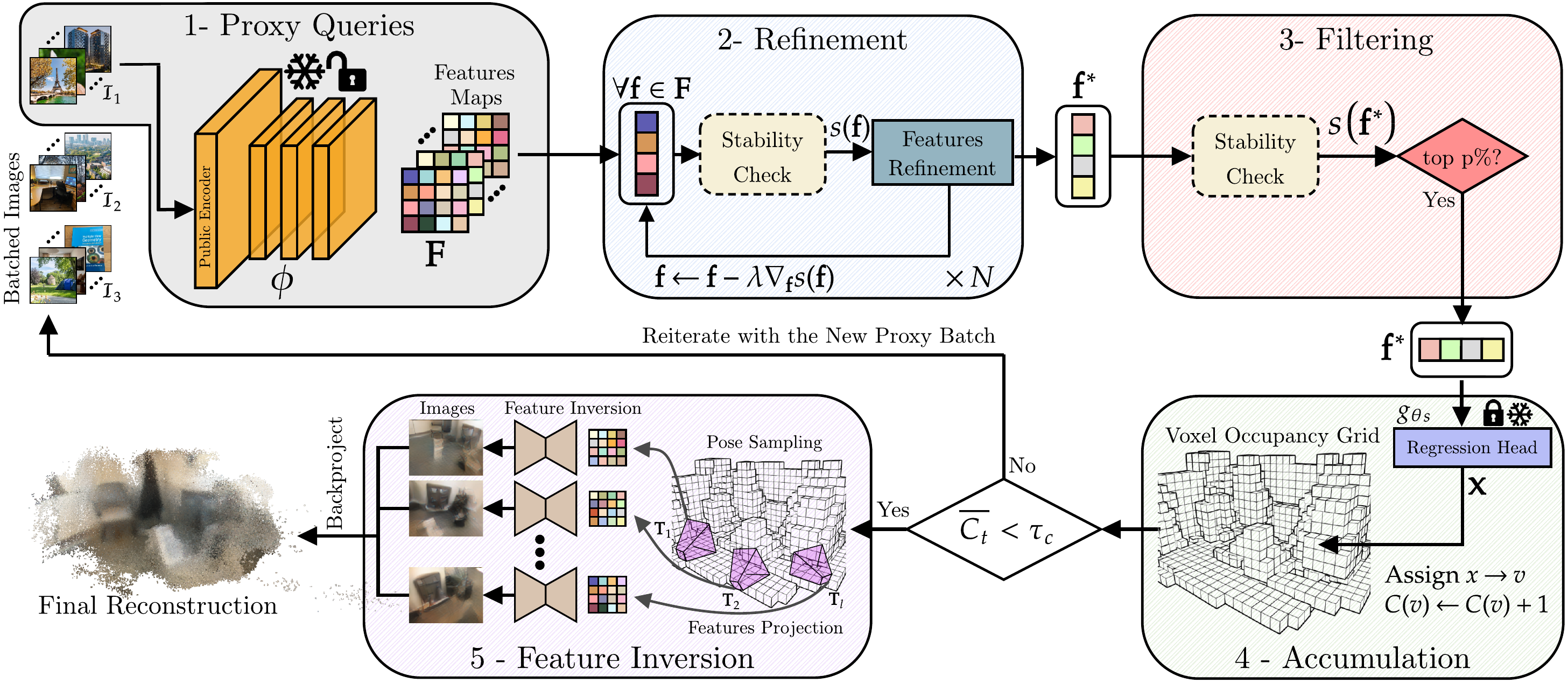}
\caption{Overview of the proposed white-box attack on SCR models. 
(1)~Proxy images are encoded using a public feature extractor to obtain dense 
feature maps. 
(2)~Features are optionally refined via gradient descent to move them toward 
stable regions of the learned scene. 
(3)~Unstable predictions are discarded by retaining only the top-$p\%$ most stable features. 
(4)~Stable 3D predictions are accumulated into a voxel occupancy grid to 
recover scene geometry. 
(5)~Recovered features are projected to virtual camera poses and inverted to 
synthesize novel views, revealing scene appearance.   \vspace{-0.1cm}} 
    \label{fig:schematic_attack}    
\end{figure}

Building on the threat model detailed above, we present a query-based attack that reconstructs scene geometry from a trained SCR model. 
For clarity, we describe the full pipeline in the white-box setting; the gray-box and black-box variants follow the same sequence but omit stages that require additional access, such as gradient-based refinement or descriptor-level operations.
An overview of the full pipeline is provided in \figurename~\ref{fig:schematic_attack}. The attack proceeds through the following stages: proxy queries (Section~\ref{sec:proxy-queries}); stability estimation (Section~\ref{sec:stability}); feature refinement (Section~\ref{sec:optimization}); batch filtering (Section~\ref{sec:filtering}); accumulation (Section~\ref{sec:accumulation}); and image reconstruction (Section~\ref{sec:image-reconstruction}). We describe each stage in detail in the following sections.

\subsection{Proxy Queries}
\label{sec:proxy-queries}

At each iteration, we sample a batch of $B$ proxy images 
$\mathcal{I}=\{I_1,\dots,I_B\}$ from natural-image 
datasets~\cite{everingham2010pascal}, extract features using the public backbone $\mathbf{F}=\phi(\mathcal{I})$, and forward them to the scene-specific head to obtain dense 3D coordinate predictions $\mathbf{X}=g_{\theta_s}(\mathbf{F})$.

Although individual predictions are noisy, the model always outputs coordinates within the geometry of the training scene. Predicted points therefore tend to fall on or near scene surfaces rather than being uniformly distributed in space. 
Aggregating predictions across many batches already yields a coarse approximation of the scene geometry; this alone constitutes our black-box attack, requiring no access to the feature extractor or model internals. The following stages 
further improve reconstruction quality when greater access is available.

\subsection{Stability Estimation}
\label{sec:stability}

\begin{figure}[tb]
    \centering
    \begin{subfigure}[b]{0.34\textwidth}
        \centering
        \includegraphics[width=\textwidth]{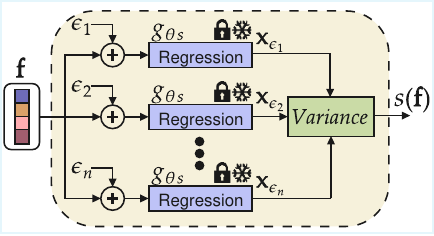}
        \caption{} %
        \label{fig:stability_module}
    \end{subfigure}
    \hfill
    \begin{subfigure}[b]{0.58\textwidth}
        \centering
        \includegraphics[width=\textwidth]{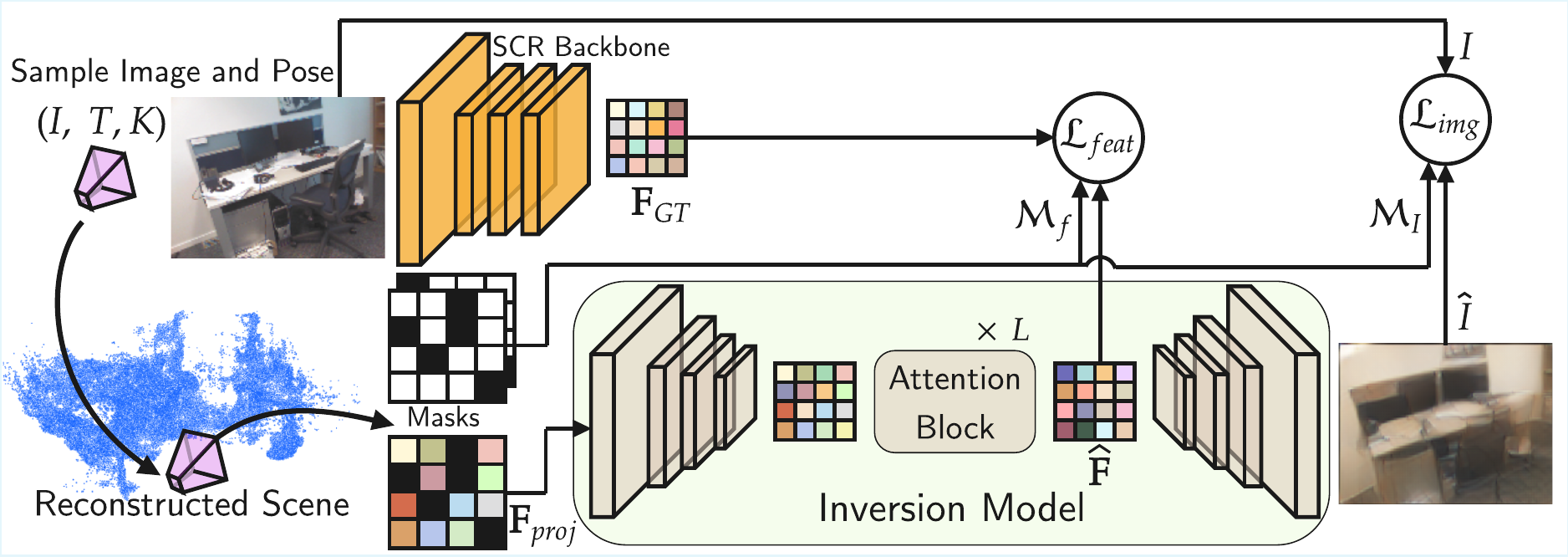}
        \caption{} %
        \label{fig:inversion_net}
    \end{subfigure}
    
    \caption{Detailed modules of the attack: (a) Stability score $s(\mathbf{f})$ computed from the variance of noise-perturbed predictions. (b) Training pipeline of the inversion network. \vspace{-0.5cm}}
    \label{fig:attack_modules}
    \vspace{-0.2cm}
\end{figure}

Proxy queries produce a dense but noisy set of 3D predictions, not all of which correspond to geometry encoded during training. Since neural networks tend to be stable near the training manifold but sensitive outside 
it~\cite{novak2018sensitivity}, we identify well-supported predictions by measuring how much the predicted 3D coordinate changes under small perturbations of its input feature.

Concretely, given a feature vector $\mathbf{f}$ extracted from a proxy image, the SCR model predicts a 3D coordinate $\mathbf{x} = g_{\theta_s}(\mathbf{f})$. We assess the stability of this prediction by perturbing the feature vector with small isotropic Gaussian noise and measuring the resulting variation in predicted 3D space, as illustrated in \figurename~\ref{fig:stability_module}. Specifically, we draw perturbations $\boldsymbol{\epsilon}_k \sim \mathcal{N}(\mathbf{0}, \sigma^2 \mathbf{I})$ and compute
\begin{equation}
s(\mathbf{f}) =
\frac{1}{n-1}\sum_{k=1}^{n}
\left\|
g_{\theta_s}(\mathbf{f}+\boldsymbol{\epsilon}_k)
-
\bar{\mathbf{x}}
\right\|_2^2,
\quad
\bar{\mathbf{x}} = \frac{1}{n}\sum_{k=1}^{n}
g_{\theta_s}(\mathbf{f}+\boldsymbol{\epsilon}_k).
\end{equation}
Low values of $s(\mathbf{f})$ indicate that $\mathbf{f}$ lies in a locally consistent region of the learned mapping; large values indicate unstable extrapolation.

Stability estimation is performed independently for each feature vector in the dense feature maps produced by a proxy batch. The resulting stability scores are later used both to refine feature representations and to filter reliable predictions.

\subsection{Feature Refinement}
\label{sec:optimization}

Proxy features are generally not aligned with those observed during training, which can lead to imprecise or inconsistent coordinate predictions. When gradients of the scene-specific head $g_{\theta_s}$ are available (white box configuration), we optionally refine the feature vectors obtained from proxy queries to encourage 
convergence toward stable regions of the learned scene. 

Concretely, given a feature vector $\mathbf{f}$, we seek a refined feature $\mathbf{f}^*$ that minimizes the stability score $s(\mathbf{f})$ defined in Section~\ref{sec:stability}:
\begin{equation}
\mathbf{f}^{*} \;=\; \arg\min_{\mathbf{f}} \; s(\mathbf{f}).
\end{equation}
This is optimized by gradient descent via updates $\mathbf{f} \leftarrow \mathbf{f} - \lambda \nabla_{\mathbf{f}}\,s(\mathbf{f})$, run for $N$ iterations or until convergence. Refinement is applied independently to each feature vector in the proxy batch.

\vspace{-0.3cm}
\subsection{Batch Filtering}
\label{sec:filtering}

After the optional refinement step (Section~\ref{sec:optimization}), some 
predictions may still lie in unstable regions and should be discarded. 
Predictions are ranked by their stability score $s(\mathbf{f})$ and only 
the top $p\%$ per batch are retained, whether applied directly to proxy 
features or to refined features $\mathbf{f}^{*}$ (depending on the access level). The resulting sparse set 
of stable 3D predictions is passed to the accumulation stage 
(Section~\ref{sec:accumulation}).

\subsection{Accumulation Across Proxy Batches}
\label{sec:accumulation}

While stability filtering acts as a local selection within each batch, accumulation serves as a global consistency check across batches. 
Stable 3D predictions from each batch are aggregated to recover a more complete estimate of the scene geometry.

Formally, let $\mathcal{P}_t=\{\mathbf{x}_i\}_{i=1}^{M_t}$ denote the retained 3D points from proxy batch $t$, expressed in the learned scene coordinate system. We discretize space into a voxel grid $\mathcal{V}$ with resolution $\Delta$ and maintain a global occupancy count $C(v)$ for each voxel $v\in\mathcal{V}$, where each point $\mathbf{x}_i$ increments the count of its corresponding voxel. Voxels consistently populated across independent batches accumulate higher counts, while spurious predictions remain weakly supported.

Accumulation continues until the average occupancy over non-empty voxels $\bar{C}_t$ falls below  a threshold $\tau_C$ or stabilizes across successive batches. 
The final reconstruction retains voxels satisfying $C(v)\geq\kappa$, where $\kappa$ is a fixed minimum occupancy threshold. 

For gray-box and white-box attacks, where stability scores are available, the most stable 3D prediction within each retained voxel is kept as its representative point. In the black-box setting, where no filtering or refinement is performed, the centroid of all predictions falling in each voxel is used instead.

\subsection{Image Reconstruction via Feature Inversion}
\label{sec:image-reconstruction}

While previous stages reconstruct scene geometry, point clouds alone do not always reveal sensitive visual details. To expose such information, we recover scene appearance by synthesizing images from arbitrary viewpoints via a feature inversion network.
Each pixel receives the feature of the nearest projected point in depth, forming a projected feature tensor $\mathbf{F}_{\text{proj}} \in \mathbb{R}^{C \times H \times W}$, with empty pixels set to zero. A CNN encoder downsamples $\mathbf{F}_{\text{proj}}$ to match the SCR feature resolution $\mathbb{R}^{C \times H' \times W'}$.
Because $\mathbf{F}_{\text{proj}}$ aggregates descriptors induced by many proxy queries, it is noisy and differs from features extracted from real images. The inversion network first refines this tensor into an adapted feature map $\widehat{\mathbf{F}}$ using a Transformer encoder with five attention blocks, then reconstructs an image $\widehat{\mathbf{I}}$ through a convolutional upsampling decoder producing an output of size $3 \times H \times W$.
Similar image reconstruction attacks have been studied for SfM maps~\cite{speciale2019privacy, dosovitskiy2016inverting, MahendranVedaldi_CVPR15}, where accurate geometry and features from the original images are available. In our setting, the geometry is recovered by the attack itself and the descriptors arise from proxy queries rather than target images, making inversion significantly more complex.
To make inversion feasible under these conditions, we train the inversion model on data that reproduces the attack setting. ACE heads are first trained on 120 ScanNet scenes and queried with the real training images to obtain ground-truth 3D points. They are then queried with proxy images from SUN RGB-D~\cite{song2015sun} to extract candidate point–descriptor pairs. To improve reliability, we retain only pairs whose predicted 3D locations lie within 4\,cm of the corresponding ground-truth points and for which at least four observations fall into the same voxel. Training samples are generated by projecting the filtered reconstructed points into target views and supervising the network with the corresponding ScanNet images.

Optimization combines an image reconstruction loss $\mathcal{L}_{img}$ (MSE + LPIPS) between $\widehat{\mathbf{I}}$ and $\mathbf{I}$ with a feature consistency loss $\mathcal{L}_{feat}$ aligning $\widehat{\mathbf{F}}$ to SCR backbone features $\mathbf{F}_{\text{gt}}$. Both losses are masked to account for sparse projections, using $\mathcal{M}_f$ for valid feature locations and a dilated mask $\mathcal{M}_I$ for image regions where projected points are present. 
 This masking is necessary because some regions, particularly scene boundaries, may be observed only few times in the training set, and low-texture areas are inherently harder to recover reliably.
The whole training pipeline and architecture is visible in \figurename~\ref{fig:inversion_net}.
Note that, using a single scene per batch leads to overfitting and strong color bias; we mitigate this effect with mixed-scene batching and random projection dropout.

At inference time, the inversion model operates on reconstructed point clouds produced by the attack, with descriptors filtered and voxelized using the same procedure as during training.%

%% file: sec/5-experiments.tex
\section{Experiments}

In this section, we evaluate the proposed attack on indoor and outdoor benchmarks across multiple architectures, attacker access levels, and query budgets.

\subsection{Experimental Setup and Implementation Details}

\noindent\textbf{Datasets.} 
We evaluate on 7-Scenes~\cite{shotton2013scene} and Cambridge Landmarks~\cite{kendall2015posenet}, covering indoor and outdoor environments. Proxy queries are drawn from PASCAL VOC~\cite{everingham2010pascal}. The feature inversion network is trained on ScanNet~\cite{dai2017scannet}.

\noindent\textbf{Implementation Details.} We use $n=4$ perturbations with 
$\sigma=0.01$, batch size $B=32$, and $N=30$ refinement iterations, retaining 
the top $p=20\%$ of predictions per batch. Voxel resolution is $\Delta=2$\,cm 
for indoor and $10$\,cm for outdoor scenes; other hyperparameters are provided in the supplementary. The inversion model is trained for 10,000 
iterations on an RTX 4090.

\noindent\textbf{Metrics.} Reconstruction quality is measured against a pseudo-GT obtained by querying the model with training images. We report F1 at 2/5\,cm (indoor) and 10/30\,cm (outdoor), where higher values indicate better reconstruction accuracy, and half Chamfer Distance (CD), where lower values indicate better geometric alignment. Image quality is reported as PSNR with higher values corresponding to better image fidelity. 
We use pseudo-GT as the reference because it reflects the geometry encoded by the model: an attack cannot recover geometry that the model itself has not learned. This also provides a model-specific reference, since different SCR architectures may encode slightly different subsets of the scene. A detailed explanation, together with visual and quantitative examples and comparisons against SfM and RGB-D ground truth, is provided in the supplementary material.
%


\subsection{Access Level vs. Reconstruction Quality}

Table~\ref{tab:attack_progression_real} reports reconstruction quality on 7-Scenes for ACE across the three attack variants. Reconstruction quality improves consistently with access level, with white-box yielding the best results across all scenes. Notably, the black-box variant already achieves competitive performance, with an average F1 of 62.3 at 2\,cm and 82.8 at 5\,cm, and gray-box scores remain close to white-box, confirming that meaningful 3D reconstruction is achievable even without access to model internals or gradients.
Stairs is the most challenging scene across all variants due to repetitive structures and textureless surfaces, which increase prediction ambiguity in the SCR model and, by extension, reduce attack efficacy. 
Qualitative results in the supplementary material further confirm this trend, showing progressively cleaner reconstructions as access increases.

%

\input{tab/attack_progression}
\vspace{-0.5cm}

\input{tab/scr_generalizatoin}
\vspace{-0.0cm}
\subsection{Generalization Across SCR Architectures}
Table~\ref{tab:cross_architecture} reports white-box results across SCR architectures on indoor and outdoor datasets.
Results obtained with the attack follow similar patterns across architectures, indicating that the leakage is not architecture-specific. %

Across both datasets, the relative behavior of the evaluated models is largely consistent. ACE and GLACE yield the most complete reconstructions, while DSAC* and ACE-G produce weaker reconstructions with noisier geometry.
\\
Reconstruction quality is overall higher on 7-Scenes than on Cambridge Landmarks. Indoor scenes are spatially smaller and geometrically simpler, which makes their structure easier to recover. In contrast, the larger spatial extent and higher variability of outdoor scenes reduce reconstruction completeness, leading to a noticeable performance drop for all models on Cambridge.
\\
Architectural differences nevertheless influence the severity of leakage. ACE appears most exposed, likely because its scene-specific MLP fits the training scene very closely, concentrating geometric information in a compact parameter space that is easier to extract. 
GLACE conditions its regression head on a global scene descriptor, which could, in principle, provide additional protection. On small indoor scenes such as 7-Scenes, where views are highly similar and spatially limited, this conditioning offers little resistance. On larger outdoor scenes, however, GLACE shows noticeably stronger resistance compared to ACE. In our experiments, the global descriptor is simply set to zero (see supplementary).
DSAC* shows comparatively stronger resistance. Unlike ACE-based models, it is trained end-to-end and does not rely on pretrained features. As a result, we believe that proxy images from unrelated datasets might produce more unstable predictions that do not consistently align with the scene geometry, leading to noisier reconstructions. Finally, ACE-G, designed to improve generalization, also reduces leakage to some extent.
Although resistance varies across architectures, substantial portions of the scene remain recoverable in all cases.

\begin{figure*}[t]
\centering
\begin{subfigure}{0.48\textwidth}
    \centering
    \includegraphics[width=\linewidth]{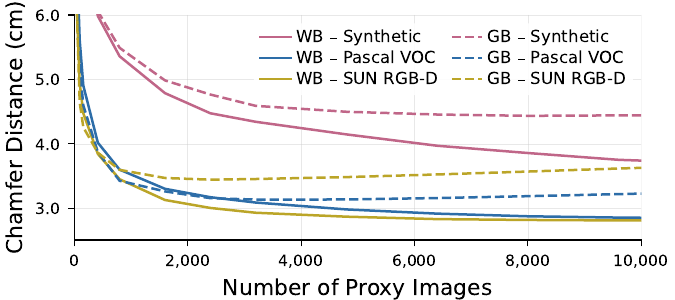}
    \includegraphics[width=\linewidth]{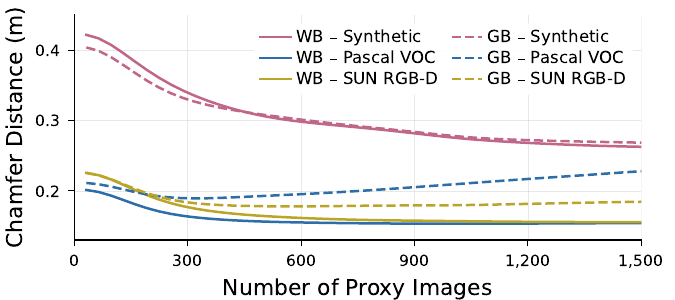}
    \caption{Proxy images vs. reconstruction quality}
\end{subfigure}
\hfill
\begin{subfigure}{0.48\textwidth}
    \centering
    \includegraphics[width=\linewidth]{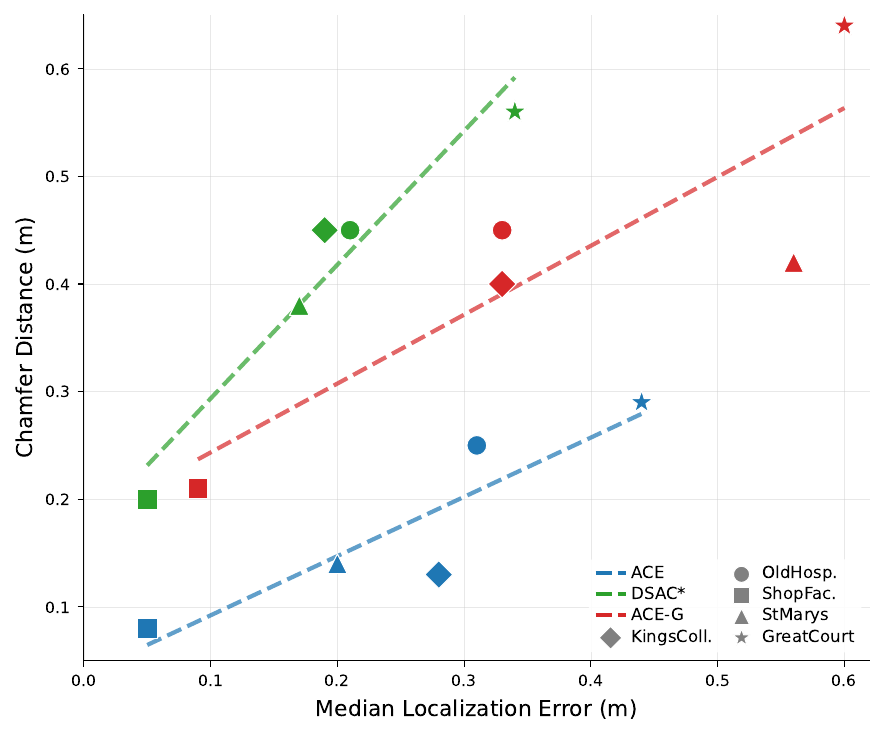}
    \caption{Reconstruction quality vs.\ model accuracy}
    \label{fig:qualvsacc}
\end{subfigure}
\caption{(a) Effect of proxy image domain on query budget for two representative scenes (top) 7-Scenes Stairs and (bottom) Cambridge Shop Facade and (b) relationship between localization model accuracy (translation) and reconstruction quality.}
\label{fig:proxy_and_attack}

\end{figure*}

We also observe that attack performance correlates with the localization accuracy of the underlying model (\figurename~\ref{fig:qualvsacc}). Models with higher localization error tend to produce less accurate reconstructions. This suggests that when the scene is not well learned for localization, its geometric structure is encoded less consistently in the network parameters, making it harder to extract. Note that GLACE is not included in this comparison, as its regression head is conditioned on a global scene descriptor, which alters this relationship.

\vspace{-0.3cm}
\subsection{Proxy Image Domains and Query Budget}
We also analyze how the domain of proxy images affects the required query budget. To study this effect, we select representative indoor and outdoor scenes: 7-Scenes Stairs and Cambridge Shop Facade; results on additional scenes are provided in the supplementary material. For these scenes, we attack ACE using white-box and gray-box variants of the attack with images drawn from three proxy sources: (1) natural images from PASCAL VOC~\cite{everingham2010pascal}, (2) indoor scene images from SUN RGB-D~\cite{song2015sun}, and (3) synthetic geometric patterns generated from random noise (see supplementary for details).

In all cases, proxy images drawn from the synthetic geometric pattern dataset lead to poor reconstruction quality, even when a large number of proxy images is used. 
In contrast, real world images produce substantially better reconstructions. For white-box attacks, the two proxy datasets eventually converge to similar reconstruction quality as the query budget increases. 
In the gray-box setting, however, we observe that the results are scene-dependent: certain scenes favor PASCAL whereas others favor SUN, highlighting the importance of proxy image selection for efficient reconstruction under limited query budgets. 
Across all sequences, using natural images, a good-quality reconstruction can be achieved with only a few hundred images with white or gray-box attacks.

\begin{figure}[tb]
    \centering
    \includegraphics[width=0.99\linewidth]{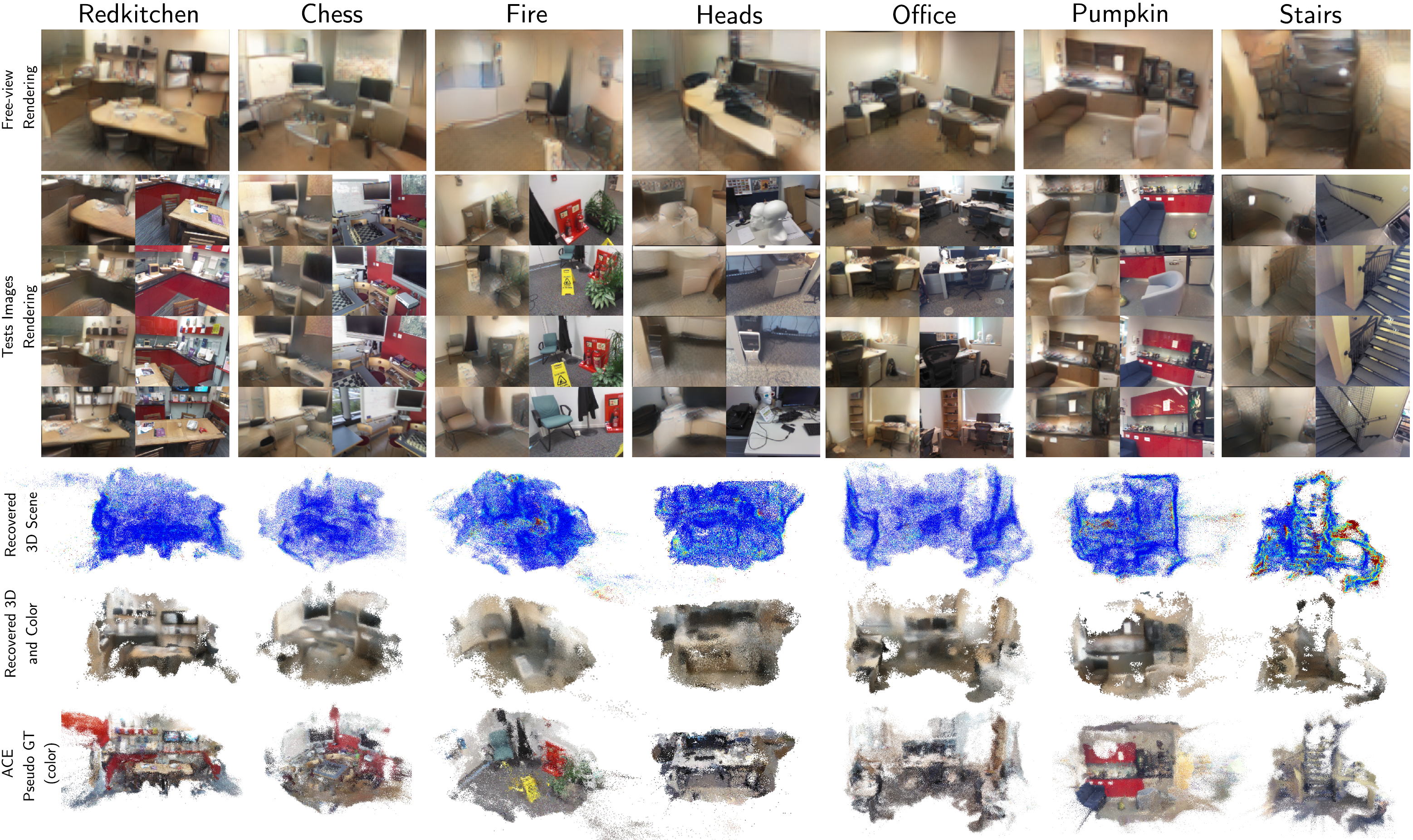}
\caption{\textbf{Qualitative reconstruction on 7-Scenes.} Free-view renderings are generated from viewpoints that are not included in either the training or test sets. Real query images are shown alongside renderings of the reconstructed scene from the same poses for visual comparison. We also show the recovered 3D geometry as an error heat map, with colors ranging from blue for 0 cm error to red for errors above 10 cm, together with the reconstructed geometry with recovered colors and the ACE pseudo-GT.}
    \label{fig:clouds_7_scenes}    
\end{figure}
    
\begin{figure}[t]
    \centering
    \includegraphics[width=0.99\linewidth]{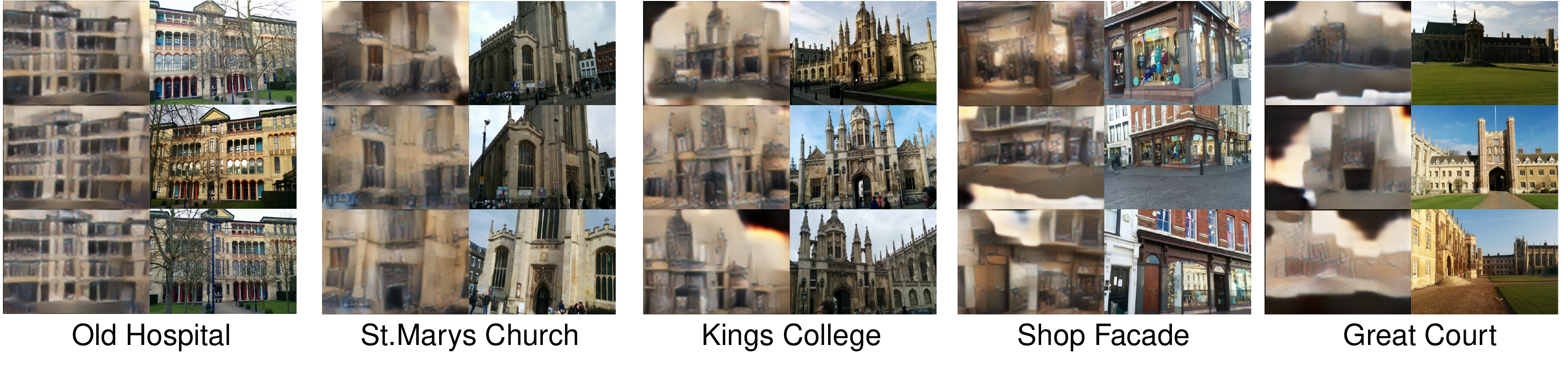}
 \caption{\textbf{Cambridge Landmarks reconstruction results.} The model reconstructs outdoor scenes despite being trained only on indoor environments. \vspace{-0.5cm}}
    \label{fig:iamge_Camb}    
\end{figure}
\subsection{Image Reconstruction}

\input{tab/psnr_scores}

We analyze the quality of the image reconstruction network by evaluating both the visual fidelity of the generated images and the consistency of the reconstructed 3D point clouds. 
Point–feature pairs are generated following the protocol in Sec.~\ref{sec:image-reconstruction}, treating the extracted points as pseudo ground truth. 
Because pairs are filtered by geometric consistency and voxel occupancy, the points used for image generation may slightly differ from those reconstructed during the attack. 
We further use the ground-truth camera poses from the original dataset to obtain reference images for comparison.

Qualitative examples for the white-box setting are shown in \figurename~\ref{fig:clouds_7_scenes}. Quantitative PSNR results are reported in Tab.~\ref{tab:psnr_attack}. The results confirm that reconstruction quality consistently improves with increased access to the underlying model.

To the best of our knowledge, this is the first work to perform image inversion directly from a trained SCR model. Despite that, we report the results of \cite{chelani2025obfuscation} as a reference, which achieves an average PSNR of 16.01 across scenes, while our best result is 15.32, where our results are also obtained under a substantially more challenging setting. We provide additional qualitative and quantitative comparisons with other PPVL inversion methods~\cite{pittaluga2019revealing}, together with an analysis of appearance leakage quality, in the supplementary material.

We attribute the remaining gap primarily to two factors. First, accurate color prediction is inherently challenging in our setting. Second, certain regions of the scene, particularly near image boundaries, are underrepresented during training, which negatively impacts reconstruction quality in those areas.

To illustrate cross-domain generalization, we additionally present qualitative results on the Cambridge dataset in \figurename~\ref{fig:iamge_Camb}. Despite being trained exclusively on indoor scenes, the model still produces meaningful predictions outdoors. %

%% file: tab/attack_progression.tex
\begin{table*}[t]
\centering
\caption{\textbf{Attack progression on ACE (7Scenes).} Comparison of Black-, Gray-, and White-box attack stages using Pascal proxy.
\textbf{Bold}: best per scene.
\vspace{-0.1cm}}
\label{tab:attack_progression_real}
\setlength{\tabcolsep}{6pt}
\resizebox{\textwidth}{!}{%
\begin{tabular}{l !{\vrule width 0.5pt} ccc !{\vrule width 0.5pt} ccc !{\vrule width 0.5pt} ccc}
\toprule
\rowcolor{gray!30}
& \multicolumn{3}{c!{\vrule width 0.5pt}}{\textbf{Black-Box}} & \multicolumn{3}{c!{\vrule width 0.5pt}}{\textbf{Gray-Box}} & \multicolumn{3}{c}{\textbf{White-Box}} \\
\rowcolor{gray!30}
& \footnotesize $F1^{2}\uparrow$ & \footnotesize $F1^{5}\uparrow$ & \footnotesize $CD\downarrow$ & \footnotesize $F1^{2}\uparrow$ & \footnotesize $F1^{5}\uparrow$ & \footnotesize $CD\downarrow$ & \footnotesize $F1^{2}\uparrow$ & \footnotesize $F1^{5}\uparrow$ & \footnotesize $CD\downarrow$ \\
\midrule
Chess      & 71.2 & 88.0 & 2.55  & 77.7 & 93.3 & 1.90  & \textbf{83.0} & \textbf{97.2} & \textbf{1.59} \\
\rowcolor{gray!10}
Fire       & 60.3 & 80.7 & 3.13  & 76.2 & 94.0 & 1.83  & \textbf{79.9} & \textbf{95.8} & \textbf{1.72} \\
Heads      & 52.0 & 77.8 & 3.03  & 70.3 & 92.6 & 1.91  & \textbf{78.1} & \textbf{95.3} & \textbf{1.78} \\
\rowcolor{gray!10}
Office     & 71.2 & 89.7 & 2.29  & 76.2 & 92.0 & 2.19  & \textbf{82.7} & \textbf{96.2} & \textbf{1.65} \\
Pumpkin    & 61.1 & 79.7 & 3.67  & 68.5 & 87.1 & 2.56  & \textbf{76.1} & \textbf{93.4} & \textbf{2.04} \\
\rowcolor{gray!10}
Redkitch.  & 74.2 & 91.5 & 2.41  & 76.5 & 92.2 & 2.08  & \textbf{82.3} & \textbf{96.5} & \textbf{1.62} \\
Stairs     & 46.2 & 72.1 & 4.79  & 48.9 & 76.1 & 3.57  & \textbf{58.8} & \textbf{84.8} & \textbf{2.96} \\
\midrule
\rowcolor{blue!10}
\textbf{Avg.} & 62.3 & 82.8 & 3.12  & 70.6 & 89.6 & 2.30  & \textbf{77.3} & \textbf{94.2} & \textbf{1.91} \\
\bottomrule
\end{tabular}%
}
\vspace{-0.2cm}
\end{table*}

%% file: tab/scr_generalizatoin.tex
\begin{table}[!ht]
\centering
\caption{\textbf{Generalization across SCR architectures.} White-box attack results on 7-Scenes (F1 at 2\,cm/5\,cm, CD in cm) and Cambridge Landmarks (F1 at 10\,cm/30\,cm, CD in m). \vspace{-0.1cm}} 
\label{tab:cross_architecture}
\setlength{\tabcolsep}{4pt} 
\resizebox{\textwidth}{!}{%
\begin{tabular}{l l !{\vrule width 0.5pt} ccc | ccc | ccc | ccc}
\toprule
\rowcolor{gray!30}
\cellcolor{gray!30} & \cellcolor{gray!30} & \multicolumn{3}{c|}{\textbf{ACE \cite{brachmann2023accelerated}}} & \multicolumn{3}{c|}{\textbf{GLACE \cite{wang2024glace}}} & \multicolumn{3}{c|}{\textbf{DSAC* \cite{brachmann2017dsac}}} & \multicolumn{3}{c}{\textbf{ACE-G \cite{bruns2025ace}}} \\
\rowcolor{gray!30}
\cellcolor{gray!30} & \cellcolor{gray!30} & \footnotesize $F1^{2}\uparrow$ & \footnotesize $F1^{5}\uparrow$ & \footnotesize $CD\downarrow$ & \footnotesize $F1^{2}\uparrow$ & \footnotesize $F1^{5}\uparrow$ & \footnotesize $CD\downarrow$ & \footnotesize $F1^{2}\uparrow$ & \footnotesize $F1^{5}\uparrow$ & \footnotesize $CD\downarrow$ & \footnotesize $F1^{2}\uparrow$ & \footnotesize $F1^{5}\uparrow$ & \footnotesize $CD\downarrow$ \\
\midrule
\cellcolor{white} & Chess      & 83.0 & 97.2 & 1.59 & 78.2 & 95.7 & 1.70 & 71.8 & 90.8 & 2.25 & 57.1 & 85.3 & 3.00 \\
\rowcolor{gray!10}
\cellcolor{gray!10} & Fire       & 79.9 & 95.8 & 1.72 & 80.4 & 96.4 & 1.67 & 71.8 & 91.4 & 2.15 & 58.9 & 85.6 & 3.19 \\
\cellcolor{white} & Heads      & 78.1 & 95.3 & 1.78 & 75.6 & 93.4 & 1.86 & 68.0 & 91.1 & 2.40 & 59.0 & 88.0 & 2.68 \\
\rowcolor{gray!10}
\cellcolor{gray!10} & Office     & 82.7 & 96.2 & 1.65 & 84.0 & 97.6 & 1.47 & 69.5 & 89.3 & 2.47 & 58.0 & 84.9 & 3.72 \\
\cellcolor{white} & Pumpkin    & 76.1 & 93.4 & 2.04 & 72.7 & 92.3 & 2.13 & 58.6 & 82.7 & 3.40 & 46.1 & 78.6 & 4.55 \\
\rowcolor{gray!10}
\cellcolor{gray!10} & Redkitch.  & 82.3 & 96.5 & 1.62 & 82.9 & 97.0 & 1.54  & 65.6 & 86.5 & 2.74 & 55.2 & 84.2 & 3.36 \\
\cellcolor{white} & Stairs     & 58.8 & 84.8 & 2.96 & 61.7 & 86.8 & 2.64 & 40.1 & 66.8 & 6.05 & 33.4 & 68.0 & 5.29 \\
\rowcolor{blue!10}
\multirow{-3}{*}[4.2em]{\rotatebox[origin=c]{90}{\textbf{7-Scenes}}} & \textbf{Avg.} & \textbf{77.3} & \textbf{94.2} & 1.91 & 76.5 & \textbf{94.2} & \textbf{1.86} & 63.6 & 85.5 & 3.07 & 52.5 & 82.1 & 3.68 \\
\midrule
\rowcolor{gray!30}
\cellcolor{gray!30} & \cellcolor{gray!30} & \footnotesize $F1^{10}\uparrow$ & \footnotesize $F1^{30}\uparrow$ & \footnotesize $CD\downarrow$ & \footnotesize $F1^{10}\uparrow$ & \footnotesize $F1^{30}\uparrow$ & \footnotesize $CD\downarrow$ & \footnotesize $F1^{10}\uparrow$ & \footnotesize $F1^{30}\uparrow$ & \footnotesize $CD\downarrow$ & \footnotesize $F1^{10}\uparrow$ & \footnotesize $F1^{30}\uparrow$ & \footnotesize $CD\downarrow$ \\
\midrule
\cellcolor{white} & KingsColl. & 64.0 & 92.5 & 0.13 & 39.0 & 73.9 & 0.25 & 22.5 & 57.1 & 0.45 & 25.1 & 62.3 & 0.40 \\
\rowcolor{gray!10}
\cellcolor{gray!10} & OldHosp.   & 45.1 & 74.8 & 0.25 & 49.6 & 77.4 & 0.24 & 29.5 & 61.7 & 0.45 & 27.2 & 69.0 & 0.45 \\
\cellcolor{white} & ShopFac.   & 83.5 & 98.8 & 0.08 & 58.2 & 86.3 & 0.16 & 50.6 & 84.9 & 0.20 & 39.5 & 80.6 & 0.21 \\
\rowcolor{gray!10}
\cellcolor{gray!10} & StMarys    & 65.1 & 91.5 & 0.14 & 16.6 & 61.6 & 0.52 & 31.2 & 67.6 & 0.38 & 14.3 & 46.9 & 0.42 \\
\cellcolor{white} & GreatCourt & 43.3 & 74.8 & 0.29 & 9.0 & 26.9 & 1.36 & 33.9 & 63.7 & 0.56 & 22.2 & 57.7 & 0.64 \\
\rowcolor{blue!10}
\multirow{-2.6}{*}[3.8em]{\rotatebox[origin=c]{90}{\textbf{Cambridge}}} & \textbf{Avg.} &  \textbf{59.0} & \textbf{85.9} & \textbf{0.18} & 34.5 & 65.2 & 0.51 & 33.5 & 67.0 & 0.41 & 25.7 & 63.3 & 0.42  \\
\bottomrule
\end{tabular}%
}
\end{table}

%% file: tab/psnr_scores.tex
\begin{table}[t]
\centering
\caption{\textbf{PSNR comparison across attack settings on ACE (7Scenes).} \vspace{-0.1cm}}
\label{tab:psnr_attack}
\setlength{\tabcolsep}{4pt}
\resizebox{0.99\linewidth}{!}{%
\begin{tabular}{l !{\vrule width 0.5pt} ccccccc >{\columncolor{blue!10}}c}
\toprule
\rowcolor{gray!30}
 & \textbf{Chess} & \textbf{Fire} & \textbf{Heads} & \textbf{Office} & \textbf{Pumpkin} & \textbf{Redkitch.} & \textbf{Stairs} & \textbf{Avg.} \\
\midrule
Black & 14.83 & 14.71 & 12.58 & 15.26 & 16.59 & 14.99 & 16.02 & 15.00 \\

Gray  & \underline{15.19} & \underline{14.72} & \underline{12.98} & \underline{15.76} & \underline{16.70} & \underline{15.14} & \underline{16.05} & \underline{15.22} \\

White & \textbf{15.29} & \textbf{14.96} & \textbf{13.01} & \textbf{15.86} & \textbf{16.72} & \textbf{15.23} & \textbf{16.15} & \textbf{15.32} \\
\bottomrule
\end{tabular}%
}
\vspace{-0.3cm}
\end{table}

%% file: sec/6-conclusion.tex
\section{Conclusion}

We presented the first attack capable of reconstructing the scene representation encoded by SCR models. By issuing proxy queries and aggregating the returned 3D predictions, the proposed approach recovers both geometric structure and appearance cues. Our method operates under multiple levels of access and reveals a previously underexplored privacy risk in visual localization systems. 

In future work, we plan to investigate more query-efficient reconstruction strategies to reduce the number of model interactions required for successful recovery.
We also aim to extend the analysis to more restrictive deployment settings, including black-box localization APIs that return only the final camera pose rather than intermediate scene-coordinate predictions. Finally, we plan to explore defense mechanisms that mitigate information leakage, including output perturbation, access control policies, and training-time regularization to limit the exposure of sensitive scene representations.

\section{Acknowledgments}

This work was supported by the Institute of Information \& Communications Technology Planning \& Evaluation(IITP)-Innovative Human Resource Development for Local Intellectualization program grant funded by the Korea government(MSIT) (IITP-2026-RS-2023-00259678).

%% file: supplementary/supl_arxiv.tex
\begingroup
\centering
{\Large\bfseries Seeing Through the Weights: Privacy Leakage in\\
Scene Coordinate Regression\par}
\vspace{0.35em}
{\small\bfseries Supplementary Material\par}
\endgroup
\vspace{1em}

\appendix
\renewcommand{\thesection}{\Alph{section}.}

\section{Qualitative Results vs. Access Level}

\begin{figure}[t]
    \centering
    \includegraphics[width=1\linewidth]{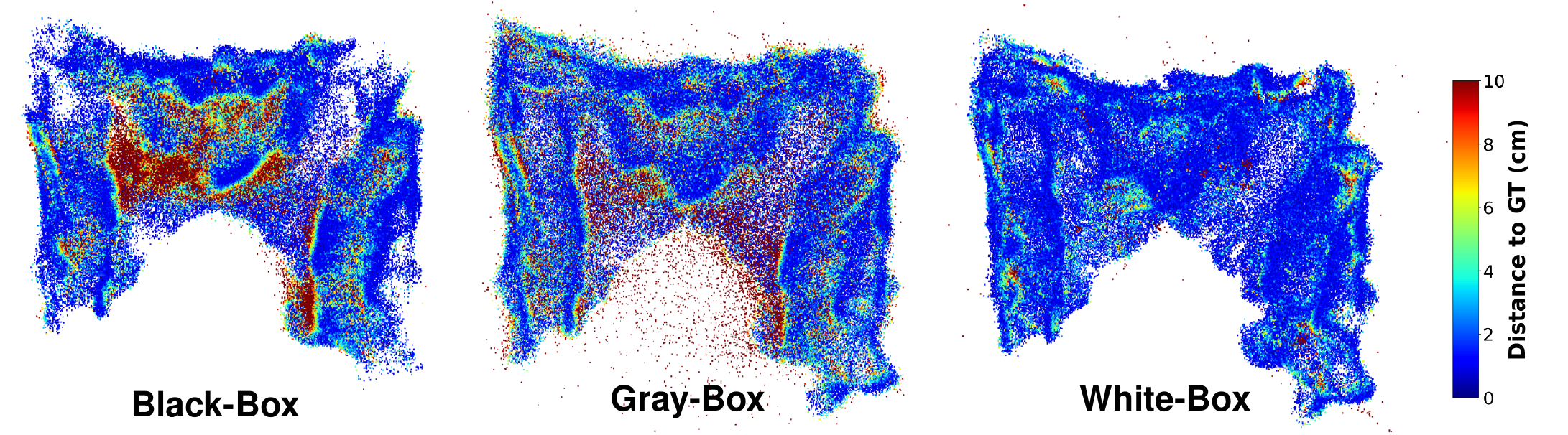}
    \caption{ Qualitative comparison of reconstructed point clouds under different access levels. Colors indicate the distance to the ground-truth point cloud (blue: small error, red: large error $\geq 10$\,cm). }
    \label{fig:comp_clouds_supss}
\end{figure}

We compare the quality of the produced point clouds under different access levels: White-box, Gray-box, and Black-box setting, generated on the Office scene from the 7-Scenes dataset (see \figurename~\ref{fig:comp_clouds_supss}).  
The figure shows a top view of the reconstructed point clouds, where the color of each point represents its distance to the ground-truth points. Blue indicates points with small distances (inliers), while red corresponds to clear outliers with distances of 10\,cm or greater. Colors between blue and red represent intermediate distance values.
The Black-box reconstruction contains noticeably fewer points overall and exhibits the highest number of outliers. 
 A visual comparison of inliers between the Gray-box and Black-box approaches is more challenging; however, both cover a large portion of the scene without noticeable gaps. In contrast, the White-box approach contains significantly fewer outliers and produces a cleaner point cloud.

Additionally, we provide image reconstruction results using these points (see \figurename~\ref{fig:comp_images_sup}). Although the overall appearance is similar across the three settings, the Black-box reconstructions exhibit noticeably lower levels of detail compared to the other two methods. Gray-box points provide slightly better reconstructions, as they capture more objects and structural elements. The difference between the White-box and Gray-box settings is less pronounced; however, the White-box results still preserve slightly finer details, and straight structures (e.g., plinths, shelves, and table contours) appear more accurately reconstructed.

\section{Proxy Image Domain type}  
In Fig.~5 of the main paper, we evaluate the robustness of our approach under domain shift by reporting attack performance using three different datasets: Pascal VOC~\cite{everingham2010pascal}, SUN RGB-D~\cite{song2015sun}, and a synthetic dataset of geometric patterns generated from random noise. These datasets represent different levels of domain discrepancy with respect to the target domain.

\begin{figure}[t]
    \centering
    \begin{subfigure}[b]{0.25\linewidth}
        \centering
        \includegraphics[width=\linewidth]{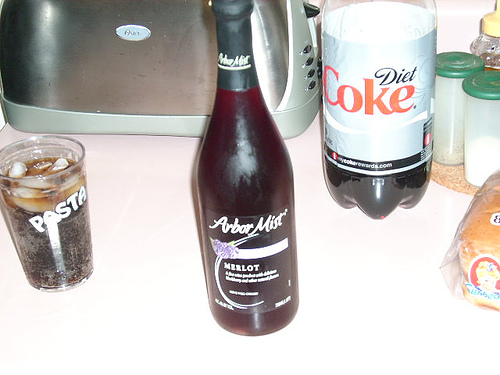}
        \caption{Pascal VOC}
    \end{subfigure}
    \hspace{0.03\linewidth}
    \begin{subfigure}[b]{0.25\linewidth}
        \centering
        \includegraphics[width=\linewidth]{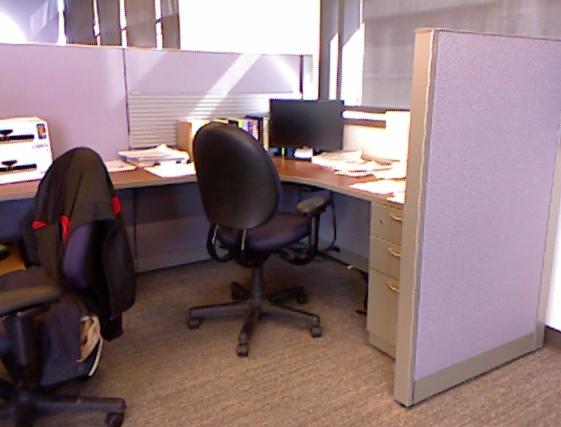}
        \caption{SUN RGB-D}
    \end{subfigure}
    \hspace{0.03\linewidth}
    \begin{subfigure}[b]{0.25\linewidth}
        \centering
        \includegraphics[width=\linewidth]{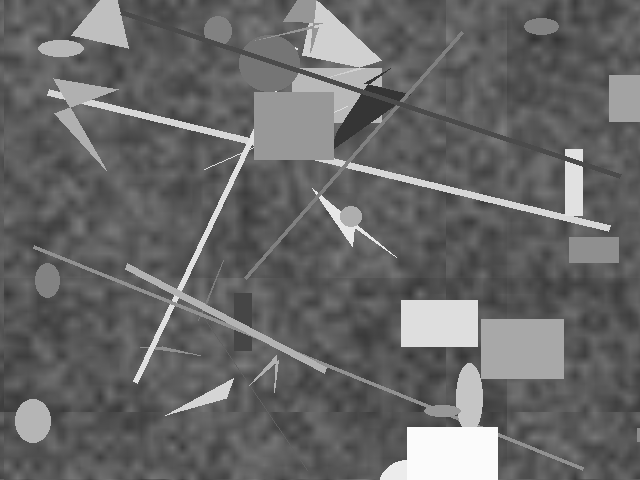}
        \caption{Synthetic Images}
    \end{subfigure}

    \caption{Sample images from different proxy datasets.}
    \label{fig:dataset_samples}
\end{figure}

SUN RGB-D is the closest to the target domain, as it contains a large number of indoor images. Pascal VOC exhibits a larger domain gap, consisting primarily of images of unrelated objects from both indoor and outdoor environments, although it still contains natural images. Finally, the synthetic dataset contains non-realistic images composed of simple geometric primitives.
The synthetic images are generated by randomly placing up to 50 geometric primitives (e.g., triangles, lines, and rectangles) within the image plane. For each object, the position, orientation, and scale are sampled randomly, while the image background is generated using low-frequency patterns. The resulting images share minimal structural similarity with real-world imagery, ensuring essentially zero domain overlap with the target dataset. Example images from all datasets are shown in \figurename~\ref{fig:dataset_samples}.

We further extend the analysis from Fig.~5 of the main paper to additional Cambridge Landmarks scenes, shown in Fig.~\ref{fig:datatypes_cambridge} of supplementary materials, where similar trends can be observed. The White-box setting consistently demonstrates better convergence across nearly all scenes, whereas the Gray-box setting may diverge when accumulating a larger number of points. For most scenes, the results obtained using Pascal VOC are slightly better than those using SUN RGB-D. We attribute this to the fact that Pascal images may be somewhat closer in feature distribution to the target scenes, as the dataset also contains outdoor images.

\section{Pseudo-GT vs. physical GT}

\begin{figure}
    \centering
    \includegraphics[width=1\linewidth]{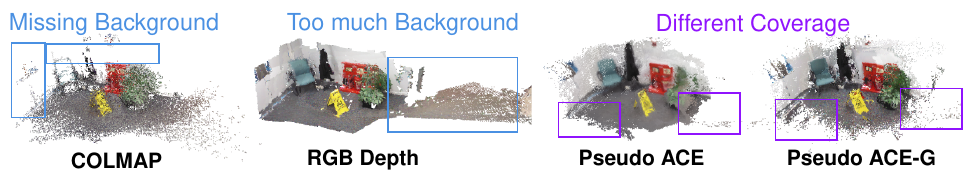}
    \caption{Colmap, depth and ACE Pseudo GT comparison on 7-scenes, scene \textit{fire}}
    \label{fig:pseudo_gt_comparison}
\end{figure}

We use pseudo-GT because it represents the geometry encoded by the model: an attack cannot recover geometry that the model did not learn. This also gives a model-specific reference, since different SCR architectures encode slightly different scene subsets (\figurename~\ref{fig:pseudo_gt_comparison}), and is consistent with SfM privacy attacks \cite{lee2023paired, moon2024efficient}, which evaluate reconstruction against the same encoded or defended SfM map. 
A unique GT point cloud is also difficult to define. COLMAP and RGB-D clouds differ significantly: SfM misses many low-texture surfaces, while RGB-D may include background geometry never encoded by the SCR model.
Following the advice of the reviewer, we report white-box scores against RGB-D and COLMAP GT in Tab.~\ref{tab:rgbd_gt}. These numbers are useful for comparison, but do not directly measure SCR leakage. \\

\input{tab/rebuttal_Pseudo_GT_small}

\vspace{-1cm}

\vspace{-1cm}



\section{Image Reconstruction Quality Evaluation}

\noindent \textbf{Upper/lower bounds.}
To jointly evaluate 3D and image reconstruction quality, we estimate lower and upper bounds for image reconstruction. For image inversion, the upper bound assumes near-perfect geometry. To simulate this, we keep only proxy-derived point-feature pairs within 2\,cm of the pseudo-GT cloud. As shown in \tableautorefname~\ref{tab:psnr_bounds}, although the upper bound is higher than the white-box PSNR score, as expected, the gap remains quite moderate, suggesting high quality of the reconstructed point cloud.

For the lower bound, we use the same pseudo-GT points, but with randomly shuffled features, to evaluate whether correct feature associations are important, or whether the model mainly relies on the correct geometry of the point cloud. In this experiment, the PSNR score drops dramatically to around 12.04 on average, which practically corresponds to nearly solid-color images that are difficult to distinguish from each other. This shows that: i) our method allows the extraction of a high-quality 3D point cloud; and ii) the trained image inversion model effectively utilizes information extracted from the backbone feature extractor.

\begin{figure}[h]
    \vspace{-0.25cm}
    \centering
    
    \includegraphics[width=0.75\linewidth]{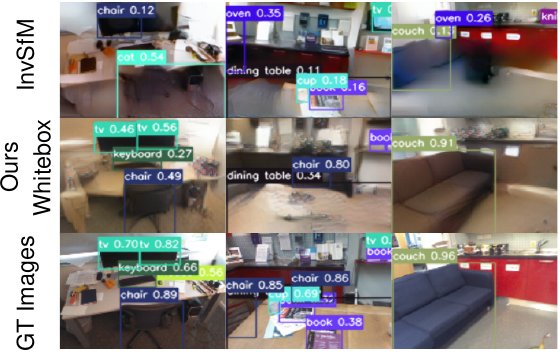}
    \caption{Example of YOLOv7 on generated images.}
    \label{fig:obj_rec}
\end{figure}

\noindent \textbf{Comparison with PPVL inversion.}
We added a comparison with InvSfM \cite{pittaluga2019revealing} (\figurename~\ref{fig:obj_rec}, \tableautorefname~\ref{tab:psnr_bounds}), which is considered an upper bound for PPVL. InvSfM preserves sharper local details when the 3D geometry is accurate, but it can still obtain low PSNR because sparse points create missing regions and occlusion artifacts.


\noindent \textbf{ Appearance leakage quality.}
To further evaluate the information revealed by the proposed method, we assess object recognizability by running YOLOv7 on the inverted images. Example results from different scenes are shown in \figureautorefname~\ref{fig:obj_rec}. Although some objects are missed or poorly reconstructed, the detector still recognizes a substantial portion of the scene content. This suggests that the recovered images preserve meaningful semantic information. Importantly, since our focus is scene understanding rather than photorealistic reconstruction, even a small number of recognizable objects can be sufficient to infer the scene context.


\begin{table}[t]
\centering
\caption{PSNR for lower/upper bound, white-box, and baselines.}
\label{tab:psnr_bounds}
\scriptsize
\setlength{\tabcolsep}{1pt}
\begin{tabular}{lcccccc}
\toprule
Metric & Lower bound & White-box & Upper bound & ACE-G & DSAC* & InvSfM w. RGB~\cite{pittaluga2019revealing}  \\
\midrule
PSNR $\uparrow$ & 12.04 & 15.32 & 15.44 & 15.37 & 15.21 & 15.48  \\
\bottomrule
\end{tabular}
\end{table}


\section{Generalization}

\begin{figure}
    \centering
    \includegraphics[width=1\linewidth]{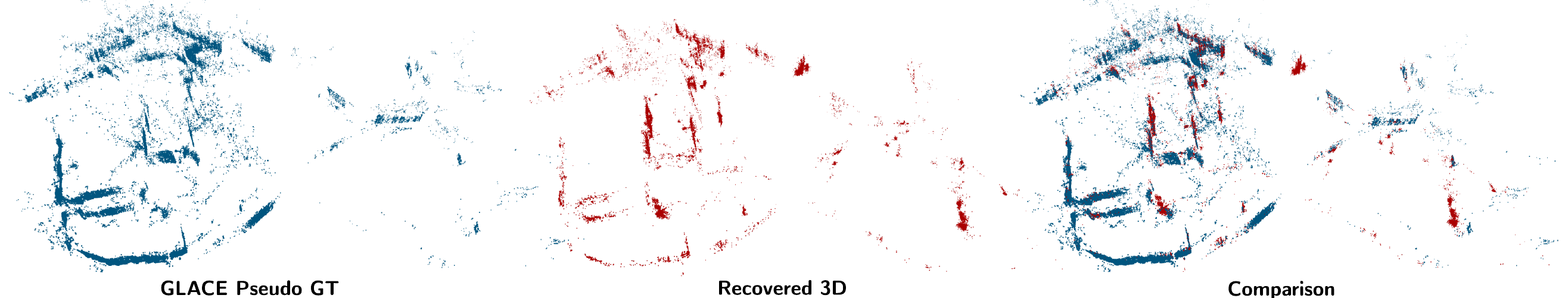}
    \caption{Whitebox attack on GLACE Aachen model.}
    \label{fig:generalization_aachen}
\end{figure}

\noindent\textbf{Large-scale reconstruction.}
Although this work mainly targets room-level indoor reconstruction, the proposed attack also generalizes to larger outdoor scenes, where both reconstruction and localization are more challenging. On GLACE Aachen, our attack recovers clear road structures and building facades, reaching 91.45\% F1@5m (\figurename~\ref{fig:generalization_aachen}). While the reconstruction remains sparse, this result confirms that SCR privacy leakage is not limited to small indoor scenes.

\begin{figure}
    \centering
    \begin{subfigure}[t]{0.55\linewidth}
        \centering
        \includegraphics[width=\linewidth]{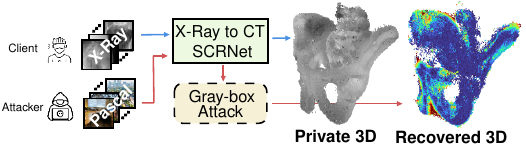}
        \caption{Overview of private and recovered 3D on SCRNet}
        \label{fig:generalization_medical}
    \end{subfigure}
    \hfill
    \begin{subfigure}[t]{0.44\linewidth}
        \centering
        \includegraphics[width=\linewidth]{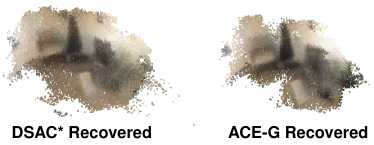}
        \caption{Recovered 3D on DSAC* and ACE-G}
        \label{fig:generalization_scr}
    \end{subfigure}
    \caption{(a) Recovered Scene Coordinates on X-ray-to-CT SCR model (b) Qualitative result of DSAC* and ACE-G.}
    \label{fig:generalization}
\end{figure}

\noindent\textbf{Medical domain.}
We also evaluate SCRNet~\cite{shrestha2023xray}, an X-ray-to-CT SCR model, using Pascal images as proxies. The gray-box attack reaches 89.96\% F1@5mm on bone geometry, showing that the leakage extends beyond natural-image scenes (\figurename~\ref{fig:generalization_medical}).

\noindent\textbf{Attack comparison on SCR Architectures}
Although the proposed method is primarily tailored to ACE-based models, which use a scene-agnostic backbone and a scene-specific MLP, we find that other SCR architectures also exhibit leakage. Scene-specific encoders, such as DSAC*, and DINO-based encoders are more robust, but they still leak recognizable geometry (\figurename~\ref{fig:generalization_scr}). Their reconstructed point clouds remain sufficiently clean for image inversion, as shown in \tableautorefname~\ref{tab:psnr_bounds}.

\vspace{-1cm}
\vspace{-1cm}



\section{Metrics}
\label{sec:supp-metrics}
Let $\mathcal{P}$ denote the reconstructed point cloud and $\mathcal{G}$ the pseudo ground-truth geometry. For each point $\mathbf{p} \in \mathcal{P}$, let $d(\mathbf{p}, \mathcal{G}) = \min_{\mathbf{g} \in \mathcal{G}} \|\mathbf{p} - \mathbf{g}\|_2$ be the nearest-neighbor distance, and define $d(\mathbf{g}, \mathcal{P})$ symmetrically.

\noindent\textbf{Chamfer Distance (CD).} We use the symmetric half Chamfer Distance as our primary geometric error metric:
\begin{equation}
\mathrm{CD}(\mathcal{P}, \mathcal{G}) =
\frac{1}{2}\left(
\frac{1}{|\mathcal{P}|}{\textstyle\sum_{\mathbf{p}\in\mathcal{P}}}
d(\mathbf{p},\mathcal{G})
+
\frac{1}{|\mathcal{G}|}{\textstyle\sum_{\mathbf{g}\in\mathcal{G}}}
d(\mathbf{g},\mathcal{P})
\right),
\end{equation}
where lower values indicate better geometric alignment.

\noindent\textbf{F1 Score.} Given a distance threshold $\delta$, and the corresponding precision $\mathrm{Prec}_\delta$ and recall $\mathrm{Rec}_\delta$, we define the \textbf{F1} score as:
\begin{equation}
    \mathrm{F1}_\delta = \frac{2 \cdot \mathrm{Prec}_\delta \cdot \mathrm{Rec}_\delta}{\mathrm{Prec}_\delta + \mathrm{Rec}_\delta}.
\end{equation}

\noindent\textbf{PSNR.} Given a predicted image $\hat{I}$ and the corresponding ground-truth image $I$, we compute the mean squared error (MSE) between them and define the \textbf{PSNR} as:
\begin{equation}
\mathrm{PSNR} = 10 \log_{10}\!\left(\frac{MAX_I^2}{\mathrm{MSE}}\right),
\end{equation}
where
\begin{equation}
\mathrm{MSE} = \frac{1}{N}\sum_{i=1}^{N}(I_i - \hat{I}_i)^2,
\end{equation}
and $MAX_I$ denotes the maximum possible pixel value, while $N$ is the number of pixels.


\section{SCR Architectures Details}
\label{sup_sec:scr_acrch_detailes}

As introduced in the main text, all evaluated models follow a common encoder-head structure: a feature encoder $\phi$ extracts dense feature maps then mapped to per-pixel 3D coordinates by a scene-specific head $g_{\theta_s}$. 
We detail the architectural specifics and attack formulation for each model below.

\noindent\textbf{ACE.}
ACE employs a pretrained frozen fully convolutional encoder that extracts 512-dimensional feature maps at $8\times$ spatial subsampling. 
The scene-specific head is a lightweight residual MLP implemented via $1\times1$ convolutions.
Because the separation between encoder and head is explicit, our attack directly follows the decomposition described in Sec.~5.3.

\noindent\textbf{GLACE.}
GLACE retains the ACE-style local encoder while augmenting it with a global image descriptor. 
Its head accepts a 768-dimensional input, formed by concatenating a 256-dimensional global scene descriptor to the 512-dimensional local features.
This descriptor is extracted using R2Former~\cite{zhu2023r2former}, a distilled Vision Transformer (DeiT-S) that outputs an L2-normalized vector per image.

\noindent\textbf{DSAC*.}
Unlike the ACE-based models, DSAC* is trained end-to-end together with differentiable RANSAC.
The encoder is a CNN sharing the convolutional residual structure of ACE, but optimized from scratch without weight sharing.
For the attack, we split the network after the second residual block to extract 512-dimensional features.
The head comprises a third residual block and three fully connected layers (implemented as $1\times1$ convolutions), adding a learned scene-mean offset to the output.
Because the encoder is scene-specific, proxy features from unrelated images exhibit lower compatibility with the learned mapping, likely causing the noisier reconstructions observed in Sec.~5.3.

\noindent\textbf{ACE-G.}
ACE-G replaces ACE’s scene-specific MLP head with a scene-agnostic coordinate regressor conditioned on a scene-specific map code.
It uses a pre-trained DINOv2 image encoder to extract patch embeddings, while the coordinate regressor predicts 3D scene coordinates from these patch features together with the learned map embeddings.
Concretely, the regressor is implemented as cross-attention-only transformer blocks, where image patch embeddings serve as query tokens and the map code provides the key and value tokens.
Under the white-box attack, we assume the attacker has full access to the pre-trained coordinate regressor and the optimized scene-specific map code.

We detail the architectural specifics and attack formulation for each model below (see Table~\ref{tab:scr-architectures}).

\begin{table}[h!]
\centering
\caption{Architecture details of the evaluated SCR models.}
\label{tab:scr-architectures}
\resizebox{\columnwidth}{!}{
\begin{tabular}{@{}lcccc@{}}
    \toprule
    & ACE~\cite{brachmann2023accelerated} & GLACE~\cite{wang2024glace} & DSAC*~\cite{brachmann2017dsac} & ACE-G~\cite{bruns2025ace} \\
    \midrule
    Encoder        & \multicolumn{2}{c}{Scene-agnostic CNN} & Scene-specific CNN & DINOv2 ViT-L/14 \\
    Input          & \multicolumn{3}{c}{Grayscale} & RGB \\
    Feature dim    & 512 & 512 & 512 & 1024 \\
    Head input dim & 512 & 768 (512+256) & 512 & 1024 + map code \\
    Conditioning   & --- & Global descriptor & --- & Map code \\
    Model size     & $\sim$5\,MB & $\sim$9\,MB & $\sim$28\,MB & $\sim$12\,MB \\
    Training time     & 5\,min. & 25\,min. & 15\,hr. & 5\,min. \\
	    \bottomrule
\end{tabular}}
\end{table}



\section{Implementation Details}
\label{sec:supp-implementation}

All attack variants use a batch size of $B=32$ proxy images drawn from PASCAL VOC~\cite{everingham2010pascal}, with voxel size $\Delta=2$\,cm for 7-Scenes and $\Delta=10$\,cm for Cambridge Landmarks.

\noindent\textbf{Pseudo ground truth.}
For evaluation only, we construct the pseudo ground-truth (GT) geometry $\mathcal{G}$ by querying the target model with all private training images. The resulting raw predictions are voxelized on the same grid ($\Delta$), and voxels with fewer than $\kappa=5$ assigned points are discarded.

\noindent\textbf{Black-Box.}
We query 4{,}000 proxy images and aggregate their predicted coordinates on the voxel grid. For each occupied voxel, we store the centroid of all predictions assigned to that voxel. Voxels with fewer than $\kappa=5$ assigned points are removed. Unlike the gray-box and white-box attacks, the black-box attack is single-pass and uses neither feature perturbation nor refinement.

\noindent\textbf{Gray-Box.}
For each batch, we perturb the encoder features $n=4$ times with Gaussian noise ($\sigma=0.01$) and compute the stability score $s(\mathbf{f})$ from the resulting per-pixel predictions. 
We retain only the top $p=20\%$ most stable predictions, i.e., those with the lowest values of $s(\mathbf{f})$. 
During voxel accumulation, each voxel stores the prediction with the lowest stability score among all predictions assigned to that voxel. 
For 7-Scenes, accumulation continues until the average number of newly occupied voxels over the last $t=30$ batches falls below $\tau_C=50$. 
For the Cambridge Landmarks dataset, to account for the outdoor nature of the scenes, which typically contain significantly more points (often several orders of magnitude larger), we increase the threshold to $\tau_C = 5000$ to accelerate convergence.
For the final filtering, we always retain only voxels with at least $\kappa=5$ entries. 

\noindent\textbf{White-Box.}
The white-box attack incorporates the feature refinement from Sec.~4.3.
We optimize this refinement using AdamW for $N=30$ iterations, with weight decay $10^{-4}$ and learning rate $\lambda=0.01$ for 7-Scenes and $\lambda=0.001$ for Cambridge Landmarks. 
All other components, including stability filtering, voxel accumulation, and the stopping criterion, remain identical to those used in the gray-box setting.

\section{Hyperparameter Ablation}
\label{sec:supp-hparam}

We provide additional experiments on the impact of the retention percentile $p$ for stability filtering on the produced point clouds quality.

\begin{figure}[h!]
    \centering
    \begin{subfigure}[t]{0.48\linewidth}
        \centering
        \includegraphics[width=\linewidth]{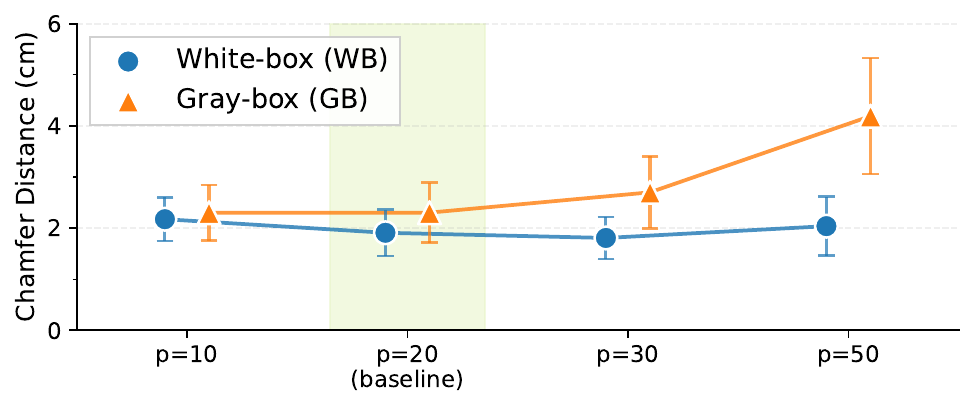}
        \caption{7-Scenes}
    \end{subfigure}
    \hfill
    \begin{subfigure}[t]{0.48\linewidth}
        \centering
        \includegraphics[width=\linewidth]{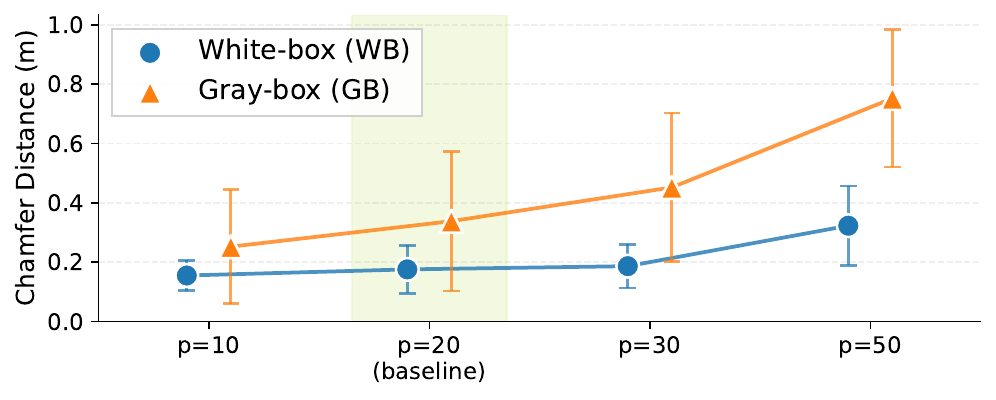}
        \caption{Cambridge Landmarks}
    \end{subfigure}
    \caption{Percentile ablation on stability-based filtering in the gray-box and white-box attacks on ACE \cite{brachmann2023accelerated}. The shaded region marks the default setting $p=20\%$, diversifying the retained percentile.}
    \label{fig:percentile_ablation}
\end{figure}

\begin{figure}[p]
    \centering
    \vspace*{\fill}
    \includegraphics[width=1\linewidth]{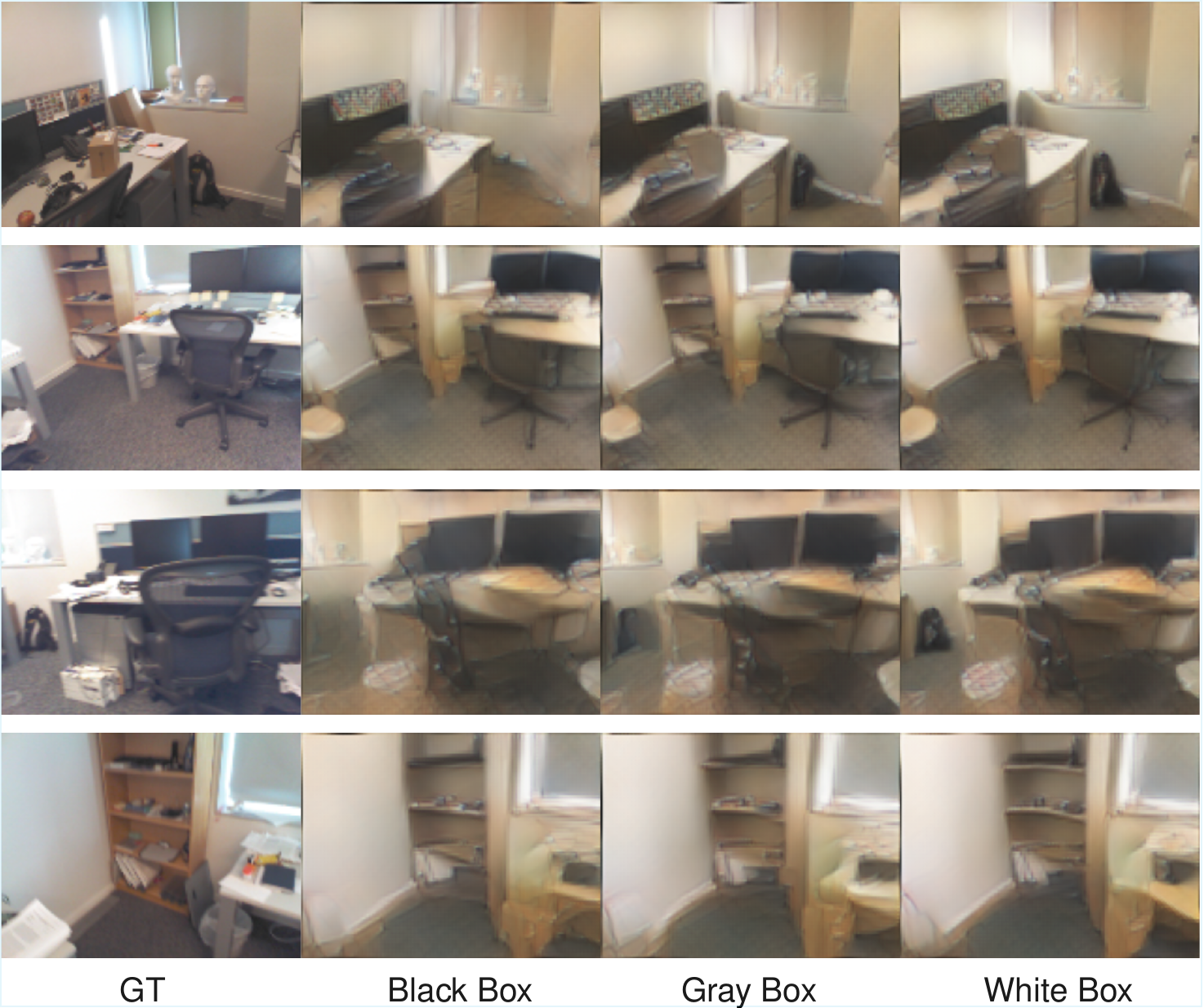}
    \caption{Comparison of image reconstruction results across different access levels on the 7-Scenes \textit{Office} scene.}
    \label{fig:comp_images_sup}
    \vspace*{\fill}
\end{figure}

\noindent
Figure~\ref{fig:percentile_ablation} evaluates the Chamfer Distance under different percentile thresholds $p \in \{10,20,30,50\}\%$ on 7-Scenes and Cambridge Landmarks. 
The white-box approach remains relatively stable across all $p$ values, with only a slight increase for $p=50\%$ on Cambridge Landmarks. This behavior results from the feature refinement step in the white-box setting, where proxy features are iteratively optimized to reduce the stability score $s(\mathbf{f})$, which drives even initially unstable features toward regions that produce consistent 3D predictions. 
In contrast, the gray-box setting does not refine features and is therefore more sensitive to $p$, with performance gradually degrading as $p$ increases for both datasets.

Importantly, even with a threshold of $p=50\%$, all configurations continue to produce meaningful reconstructions, indicating that a wide range of thresholds performs well.

\vspace{-1cm}

\section{Global Descriptors} 

 %

As explained in Sec.~\ref{sup_sec:scr_acrch_detailes}, GLACE uses global descriptors for coordinate prediction. However, since we assume white-box access and want to avoid predictions varying too strongly with the data distribution, we replace the image descriptors with zero features. This ensures a fair comparison across all datasets. 
Although better results might be achieved by optimizing the global descriptors or by using descriptors extracted from datasets with a closer domain, we leave this direction for future work.

\section{Training the Inversion Model}  
\noindent
\textbf{Input construction.}
We first project the 3D points into the image plane using the camera parameters. For each point, we compute its pixel location and depth. We then construct a projected feature tensor of size $H \times W \times D$, where each pixel stores the feature of the closest point in depth. Pixels that receive no projected points are filled with zeros.

In practice, the projected tensor is highly sparse because the learned 3D points do not densely cover the image plane. Empirically, up to 70\% of the pixels may remain empty, while model is able to produce some meaningful reconstruction. Since appearance cannot be reliably predicted in regions without geometric support, we explicitly track valid projection locations with a binary mask and use it during training. 
We denote the binary validity mask of projected features as $\mathbf{M}_f \in \{0,1\}^{H' \times W'}$, where $\mathbf{M}_f(u,v)=1$ indicates that at least one 3D point projects to pixel $(u,v)$. Here $H' \times W'$ is resolution of the feature mask. 
For image supervision, we construct a dilated image support mask $\mathbf{M}_I \in \{0,1\}^{H \times W}$ by expanding $\mathbf{M}_f$ with a fixed-radius neighborhood (10 pixels in our experiments), reflecting regions where appearance can be reasonably inferred.

\par\medskip
\noindent
\textbf{Training objectives.}
\par\smallskip
\noindent\textbf{Feature loss.}
Let $\hat{\mathbf{F}} \in \mathbb{R}^{H' \times W' \times D}$ denote the predicted feature map and $\mathbf{F}^{gt}$ the target ACE features. 
We compute the masked feature loss only at valid projection locations:


\begin{equation}
\begin{aligned}
\mathcal{L}_{feat} &=
\frac{1} {\sum_{u,v} \mathbf{M}_f(u,v) + \epsilon}\\
&\sum_{u,v} \mathbf{M}_f(u,v) \quad \left\|
\hat{\mathbf{F}}(u,v) - \mathbf{F}^{gt}(u,v)
\right\|_1
\end{aligned}
\end{equation}

\noindent\textbf{Image reconstruction loss.}
Let $\hat{\mathbf{I}}$ and $\mathbf{I}^{gt}$ denote the predicted and ground-truth RGB images. 
We apply supervision only in regions supported by projected geometry using the dilated mask $\mathbf{M}_I$:

\begin{equation}
\begin{aligned}
\mathcal{L}_{img}^{MSE} &=
\frac{1}{\sum_{u,v} \mathbf{M}_I(u,v) + \epsilon}\\
&\sum_{u,v} \mathbf{M}_I(u,v)
\left\|
\hat{\mathbf{I}}(u,v) - \mathbf{I}^{gt}(u,v)
\right\|_2^2 
\end{aligned}
\end{equation}

We additionally use a masked perceptual loss:

\begin{equation}
\mathcal{L}_{img}^{LPIPS} =
\mathrm{LPIPS}\big(
\hat{\mathbf{I}} \odot \mathbf{M}_I,\;
\mathbf{I}^{gt} \odot \mathbf{M}_I
\big),
\end{equation}

where $\odot$ denotes element-wise multiplication.

\par\medskip
\noindent\textbf{Total loss.}
\par\smallskip
\noindent
The final objective is:

\begin{equation}
\mathcal{L} =
\lambda_{img}\left(
\mathcal{L}_{img}^{MSE} +
\mathcal{L}_{img}^{LPIPS}
\right)
+
\lambda_{feat}\mathcal{L}_{feat}.
\end{equation}

Feature loss is computed only at pixels where a projected 3D point exists.
Image loss is computed within a local neighborhood (10-pixel radius) around valid projections, where appearance can be reasonably inferred.
This masking strategy prevents the network from learning from unsupported regions and stabilizes training under sparse geometric coverage.
During training we used $\lambda_{img} = \lambda_{feat}=1$.

\par\medskip
\noindent\textbf{Architecture details.}
We train a model that reconstructs images at a resolution that is $4\times$ downsampled compared to the original input. For example, this corresponds to a resolution of $160\times120$ for the 7-Scenes dataset. This resolution provides meaningful image reconstruction while preventing the model from overfitting and reducing memory requirements, allowing the model to be trained on a single RTX 4090 GPU.

For the CNN encoder, we use a sequence of convolutional blocks with $3\times3$ kernels and stride 2, followed by batch normalization and a SiLU activation layer (except for the last block). The CNN decoder consists of three convolutional blocks that gradually reduce the feature dimension from 512 to 64, followed by a series of transposed convolutions that upsample the features to the desired image resolution, with a sigmoid activation applied at the output.
For the transformer encoder, we use five layers of FlashAttention with RoPE positional encoding. A larger number of layers often led to overfitting.

\par\medskip
\noindent\textbf{Training details.}
The model is trained for 10,000 iterations with a batch size of 8, using the AdamW optimizer with an initial learning rate of $10^{-4}$. We use a CosineAnnealingLR scheduler to gradually decrease the learning rate to $10^{-6}$ by the end of training.

\section{Cost analysis and query budget}

We added runtime and memory footprint in Table~\ref{tab:runtime_memory} measured on RTX4090 with a batch size of 32. Furthermore, query statistics are available in Figure~\textbf{5} of the main manuscript and Figure~\ref{fig:datatypes_cambridge} in supplementary.
Image inversion can be run at 32 FPS.
\input{tab/rebuttal_Runtime}

\clearpage
\clearpage

\begin{figure}[p]
    \centering
    \vspace*{\fill}
    \includegraphics[width=1\linewidth]{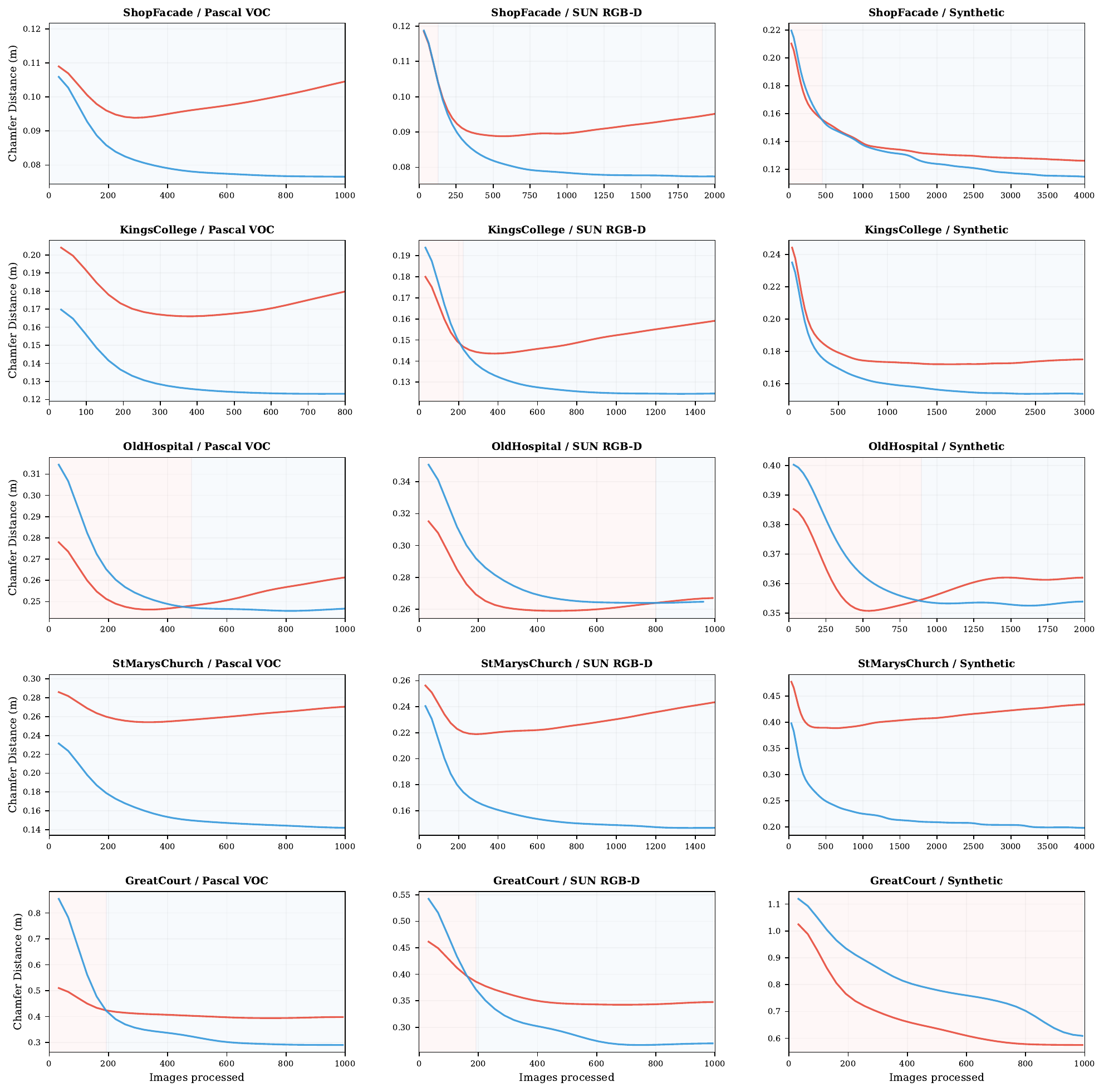}
    \caption{Gray-Box (red) vs. White-Box (blue) convergence on Cambridge Landmarks. Shaded regions indicate the better-performing attack (blue = White-Box, red = Gray-Box). White-Box achieves lower Chamfer Distance in nearly all cases as it converges, while Gray-Box often tends to diverge as more images are processed.}
    \label{fig:datatypes_cambridge}
    \vspace*{\fill}
\end{figure}
\clearpage



%% file: tab/rebuttal_Pseudo_GT_small.tex
\begin{table}[H]
\vspace{-1cm}
\centering
\caption{\textbf{Whitebox Attack vs. RGB-D and COLMAP ground truth.} Average reconstruction quality on 7-Scenes.
}
\label{tab:rgbd_gt}
\setlength{\tabcolsep}{4pt}
\resizebox{\columnwidth}{!}{%
\begin{tabular}{l|ccc|ccc|ccc|ccc}
\toprule
& \multicolumn{3}{c|}{\textbf{ACE}} & \multicolumn{3}{c|}{\textbf{GLACE}} & \multicolumn{3}{c|}{\textbf{DSAC*}} & \multicolumn{3}{c}{\textbf{ACE-G}} \\
& \footnotesize $F1^{2}\uparrow$ & \footnotesize $F1^{5}\uparrow$ & \footnotesize $CD\downarrow$ & \footnotesize $F1^{2}\uparrow$ & \footnotesize $F1^{5}\uparrow$ & \footnotesize $CD\downarrow$ & \footnotesize $F1^{2}\uparrow$ & \footnotesize $F1^{5}\uparrow$ & \footnotesize $CD\downarrow$ & \footnotesize $F1^{2}\uparrow$ & \footnotesize $F1^{5}\uparrow$ & \footnotesize $CD\downarrow$ \\
\midrule
RGB-D & 37.7 & 61.7 & 7.75 & 38.9 & 63.7 & 7.35 & 30.7 & 56.2 & 7.81 & 32.4 & 57.2 & 8.44 \\
COLMAP & 38.4 & 69.8 & 5.06 & 40.6 & 71.7 & 4.78 & 36.1 & 68.5 & 5.06 & 36.5 & 67.1 & 5.94 \\
\bottomrule
\end{tabular}%
}
\vspace{-0.2cm}
\end{table}

%% file: tab/rebuttal_Runtime.tex
\begin{table}[H]
\vspace{-0.5cm}
\centering
\caption{Mean runtime and peak GPU memory using 1K images.}
\label{tab:runtime_memory}
\setlength{\tabcolsep}{4pt}
\resizebox{\columnwidth}{!}{%
\begin{tabular}{lccccc}
\toprule
Metric & Blackbox & Graybox & Whitebox & GLACE & DSAC* \\
\midrule
Runtime & 18s $\pm$ 0s & 30s $\pm$ 1s & 4m 59s $\pm$ 6s & 9m 12s $\pm$ 3s & 3m 26s $\pm$ 17s \\
Peak GPU Mem. (GB) & 3.84 & 4.25 & 16.85 & 18.25 & 13.33 \\
\bottomrule
\end{tabular}%
}
\vspace{-0.2cm}
\end{table}